\documentclass[letterpaper, 10 pt, conference]{IEEEtran}

\IEEEoverridecommandlockouts                              
\usepackage{graphicx} % for pdf, bitmapped graphics files
\usepackage[cmyk]{xcolor}
\usepackage{epsfig} % for postscript graphics files
\usepackage{amsmath} % assumes amsmath package installed
\usepackage{amssymb}  % assumes amsmath package installed
\usepackage{bm}
\usepackage{microtype}
\usepackage{placeins}
\usepackage{siunitx}
\usepackage{subcaption}
\usepackage{units}
\usepackage{xcolor}
\usepackage{cite}
\usepackage[symbol]{footmisc}

\usepackage{textcomp}

\usepackage{multirow}
\usepackage{bigstrut}
\usepackage{booktabs}
\usepackage{pbox} % for multi line header in tables
\usepackage{makecell}
\usepackage{array}
\usepackage{tablefootnote}          % footnotes in table
\usepackage{scrextend}              % same footnote multiple times in same table

\usepackage{multirow} % multi rows in tables
\usepackage{multicol} % multicolumns in tables
\usepackage{tablefootnote}

\newcommand{\mytilde}{\raise.17ex\hbox{$\scriptstyle\mathtt{\sim}$}}
\newcommand{\myless}{\raise.27ex\hbox{\scriptsize\textless $\scriptstyle\mathtt{\sim}$}}
\newcommand{\mymore}{\raise.27ex\hbox{\scriptsize\textgreater $\scriptstyle\mathtt{\sim}$}}

\usepackage{tikz}
\usepackage{balance}
\title{Cat-like Jumping and Landing of Legged Robots in Low-gravity Using Deep Reinforcement Learning}

\author{Nikita Rudin, Hendrik Kolvenbach, Vassilios Tsounis and Marco Hutter 

\thanks{All the authors are with the Robotic Systems Lab  (RSL) at ETH Zurich, Switzerland. email: {\tt\small rudinn@ethz.ch}}%
\thanks{This work was supported by the European Space Agency (ESA) and Airbus DS in the framework of the Network Partnering Initiative 481-2016.}
}

\begin{document}
\maketitle
%\thispagestyle{empty}
%\pagestyle{empty}

%%%%%%%%%%%%%%%%%%%%%%%%%%%%%%%%%%%%%%%%%%%%%%%%%%%%%%%%%%%%%%%%%%%%%%%%%%%%%%%%
\begin{abstract}
We show that learned policies can be applied to solve legged locomotion control tasks with extensive flight phases, such as those encountered in space exploration. Using an off-the-shelf deep reinforcement learning algorithm, we trained a neural network to control a jumping quadruped robot while solely using its limbs for attitude control. We present tasks of increasing complexity leading to a combination of three-dimensional (re-)orientation and landing locomotion behaviors of a quadruped robot traversing simulated low-gravity celestial bodies. We show that our approach easily generalizes across these tasks and successfully trains policies for each case. Using sim-to-real transfer, we deploy trained policies in the real world on the SpaceBok robot placed on an experimental testbed designed for two-dimensional micro-gravity experiments. The experimental results demonstrate that repetitive, controlled jumping and landing with natural agility is possible. 

\end{abstract}


%%%%%%%%%%%%%%%%%%%%%%%%%%%%%%%%%%%%%%%%%%%%%%%%%%%%%%%%%%%%%%%%%%%%%%%
\section{INTRODUCTION}
\label{sec_introduction}
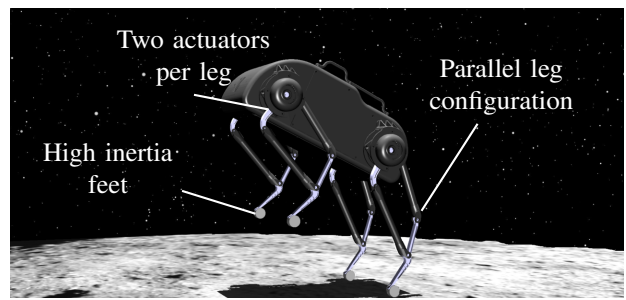
\begin{figure}[!t]
\centering
\usetikzlibrary{quotes,angles}
\begin{tikzpicture}
    \node[anchor=south west,inner sep=0] (image) at (0,0) {\includegraphics[width=0.45\textwidth, trim={0, 30mm, 0, 80mm}, clip]{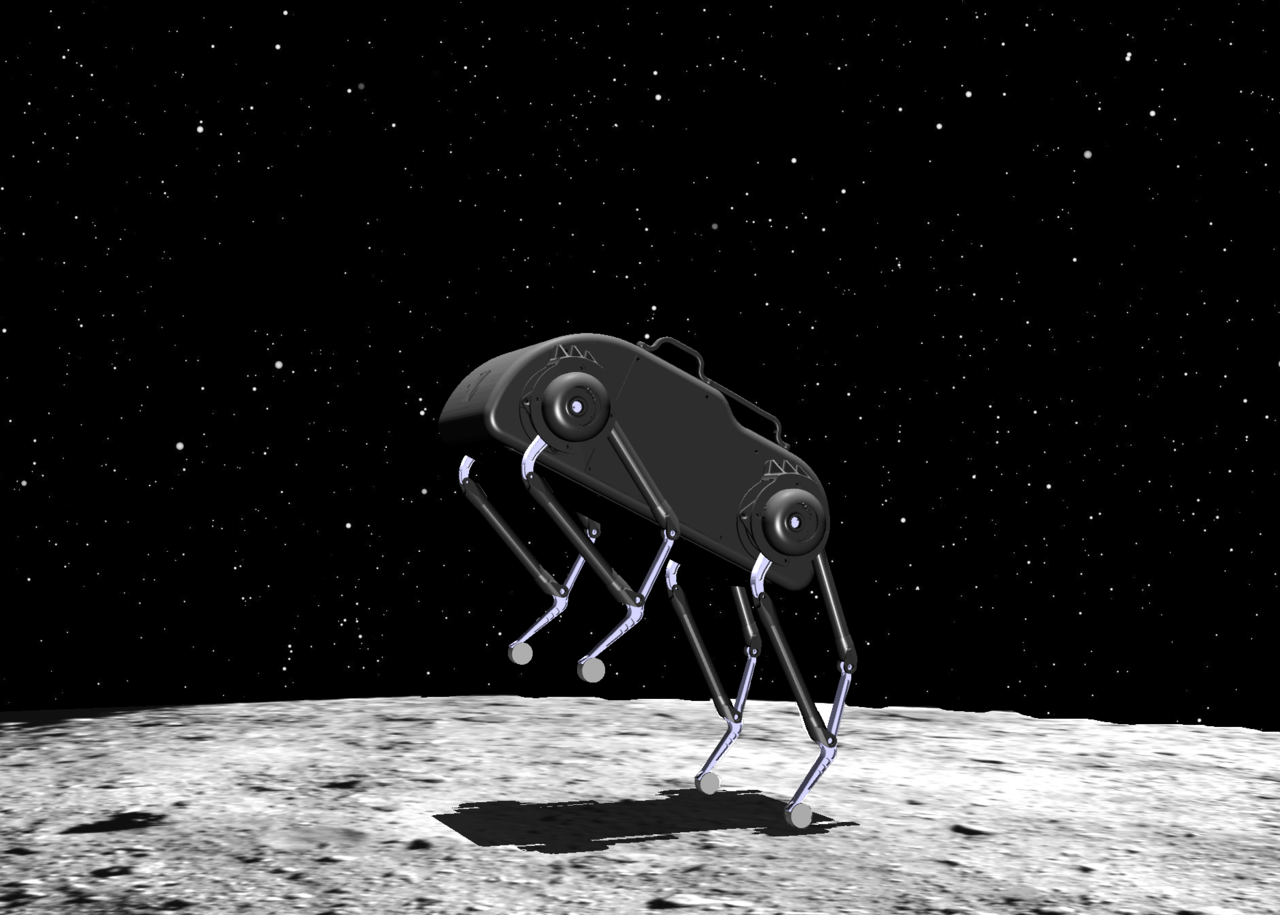}};
    \begin{scope}[x={(image.south east)},y={(image.north west)}]
        \node [anchor=south, text=white, text width=3cm,align=center] (leg) at (0.8,0.6) { Parallel leg configuration};
        \node [anchor=south, text=white,align=center] (feet) at (0.16, 0.33) {High inertia\\ feet};
        \node [anchor=south, text=white, text width=3cm,align=center] (actuators) at (0.3, 0.7) {Two actuators per leg};
        \draw [ thick, white] (leg) -- (0.67,0.3);
        \draw [ thick, white] (feet) -- (0.4,0.29);
        \draw [ thick, white] (actuators.south) -- (0.42,0.65);
        \coordinate (a) at (0.55, 0.5);
        \coordinate (b) at (0.25,0.5);
        \coordinate (c) at (0.28,0.7);
        \coordinate (d) at (0.53,0.5);
        \coordinate (e) at (0.57,0.5);
        \coordinate (f) at (0.55,0.5);
        \coordinate (g) at (0.53,0.1);
        \coordinate (h) at (0.57,0.1);
        \coordinate (i) at (0.55,0.1);
    \end{scope}
\end{tikzpicture}%
\caption{SpaceBok landing on an asteroid. The robot uses the inertia of its feet to control its orientation. The parallel leg configuration allows high torque jumps and landings, giving the feet two degrees in the vertical plane.}
\label{spacebokinspace}
\end{figure}

During recent years, legged robots have increasingly matured and demonstrated versatile, all-terrain locomotion  \cite{bellicoso2018applications}. Previous analysis has demonstrated that the advanced mobility offered by legged robots is also well suited for planetary exploration, enabling access to terrains that have so far been out of reach for wheeled rovers \cite{bartsch2012spaceclimber}. In this context, we developed the dynamically walking legged robot SpaceBok \cite{Spacebok} (Figure \ref{spacebokinspace}) for experimental validation.
Especially in low-gravity planetary environments, a robot capable of gaits with extended flight-phases poses an interesting solution for fast and efficient travel \cite{Kolvenbach2018isairas}. An extreme example of this principle was demonstrated by the Rosetta lander Philae, which performed a traversal of one kilometer in a series of unintentional ballistic flights on the surface of comet 67P/Churyumov-Gerasimenko \cite{biele2015science}.

This work focuses on the topic of controlled, low-gravity locomotion with quadruped robots.
A strong emphasis is placed on experimental verification on the physical robot.

In earth-like gravity conditions, the attitude control is mainly achieved through reaction forces with the ground. For this case, the balance control of jumping robots is well-established and has been demonstrated using inverted pendulum models in the past \cite{RaibertBalance, PoulakakisSLIP}. In our application, the problem is fundamentally different. In low-gravity conditions, jumping gaits have significantly longer flight-phases with respect to the stance-phases, which makes attitude control through ground reaction forces alone impractical. In particular, any (small) momentum obtained during take-off will result in a significant attitude error during the flight that needs to be corrected to avoid crash landings. In our previous work \cite{kolvenbach2019iros}, this observation leads to the integration of a reaction wheel, which is the standard method for controlling the attitude of satellites and other free-floating objects. Unfortunately, the working principle of reaction wheels requires them to be relatively large, heavy, and require additional actuation, making them ill-suited for agile and dynamic systems.

We observe that in nature, cats and other agile animals make extended use of their body and limbs for attitude control, which allows them to, for example, safely land back on their feet after falls \cite{kane1969dynamical}. Astronauts and cosmonauts aboard the International Space Station also learn to use their arms and legs to reorient themselves. Inspired by these observations, we develop a new approach to control SpaceBok's orientation, using only its limbs and enable precision jumping and landing.

A related problem is the attitude control of spacecrafts with robotic arms. These systems can be separated into two categories. The first category, free-flying systems can use reaction wheels and thrusters to counteract the effects of their robotic arms. In this case, it is possible to compute the torques exerted by the manipulator motion using the conservation of angular momentum and counteract it precisely \cite{YoshidaResolved}. In the second case, closer to our target application, the reaction system can not be used and the attitude of the spacecraft must be controlled solely through the motion of the arm. This is much more challenging, as it has been shown that the kinematics are non-holonomic and the dynamics have singularities \cite{PapadopoulosNonHolonomic, PapadopoulosFreeFloating}, making purely reactive control impossible. In a simplified case, an approximation of the optimal trajectory was calculated using non-linear optimization techniques \cite{SpacecraftInternalMotion}, but this approach requires extensive computation, does not scale well to multiple arms/legs, does not avoid self-collisions between the spacecraft and the manipulator, and can not handle contacts with external objects. Alternatively, it is possible to avoid complex dynamics and expensive computations by combining and repeating small cyclic motions of the arm, but this leads to a very slow re-orientation maneuver \cite{Vafa1993}. More recently, Genetic algorithms have been used to solve the task \cite{ZhangGeneticAlg}, but the proposed approach restricts the set of possible solutions, requires a new optimization for each trajectory and was only demonstrated in a simple 2D scenario with a two-link robotic arm. To the best of our knowledge, none of these methods deal with the complexity arising from the combination of the following challenges:
\begin{enumerate}
    \item A three-dimensional reorientation task using multiple legs/arms has highly non-linear dynamics. This is highlighted by the fact that advanced integration methods are needed to correctly simulate the system as described in Section \ref{sec_simulator_selection}.
    \item Self-collisions between several  limbs need to be prevented to avoid damage to the system. 
    \item The control method needs to handle discontinuities invoked by hard contacts with the ground and solve the balancing problem once contact is established.
    \item The attitude control must be highly performant. The robot must be able to re-orient itself from any configuration in a few seconds.
    \item The policy reaction must be immediate in order to prevent the robot from crashing into the ground. The limited computational budget combined with the relatively long time horizon of the re-orientation motion renders offline planning  unsuitable. 
\end{enumerate}
We decide to tackle these challenges using model-free control via Reinforcement Learning (RL), which has shown impressive results when it comes to the control of complex motions. It has been used to solve a Rubik's cube with a robotic hand \cite{RubiksCube}, learn locomotion on complex terrains, \cite{DeepmindParkour, DeepGait}, play table tennis \cite{PingPong}, teach robots to imitate animals \cite{AnimalImitation} and stand up from arbitrary initial conditions \cite{AnymalRL}. In addition to its capacity to solve complex tasks, once trained, RL has the advantage of requiring much less computation than optimization methods. The policy can be computed in a few microseconds on the on-board computer of the robot \cite{AnymalRL}.

In this work, we demonstrate our approach to tasks of increasing complexity. First, we present a two-dimensional task in which the free-floating robot must reach a given pitch angle by solely controlling its leg movements. Once deployed on the hardware, we show that the learned policy solves the task faster and more precisely than handcrafted leg motions.

We then add a landing and jumping phase. The robot must land and jump back from a wall in a given direction before performing a flip and landing on the opposite wall. This task resembles navigation in a large spacecraft or jumping on a low-gravity asteroid. The overall approach is tested and validated at the European Space Agency, where a specially designed testbed allows the simulation of two-dimensional microgravity environments.

We further increase the challenge of the task by switching to three-dimensional environments in simulation. We extend the orientation task by having the policy re-orient the free-floating robot from all possible initial orientations. Despite the limitations of our robot, which has only two degrees of freedom legs, reducing the yaw and roll axes controllability, we show that we can train a policy that controls all three axes and reaches the target from any initial state. Additionally, the simulated environment allows us to add an abduction joint to the legs. We show that in this case, a learned policy controls all axis with high performance.

Finally, our most challenging task closely resembles landing on an asteroid. The robot starts from a random orientation flying towards the ground of a low gravity body\footnote{We choose to use the gravity of Ceres, $-0.27 m/s^2 \approx 0.03g$ \cite{Ceres}. Ceres is the smallest dwarf planet of the solar system. The low gravity environment makes it an interesting target for jumping locomotion.}. It has a few seconds to re-orient itself to land on its feet softly. We choose a relatively low gravity to have a long flight giving the robot enough time to perform a complete flip while still having a manageable velocity on impact. On the other hand, the low gravity makes it harder to land softly without bouncing.

This paper is structured as follows: We present our approach in Section \ref{sec_method} and each task, with results from simulation and deployment on the physical robot, in Sections [\ref{sec_2d_attitude_control}-\ref{sec_3d_land}] respectively. Finally, we discuss the work in Section \ref{sec_discussion} and draw a conclusion in Section \ref{sec_conclusion}. A supplementary video\footnote{https://youtu.be/KQhlZa42fe4} presents our experimental results.

%%%%%%%%%%%%%%%%%%%%%%%%%%%%%%%%%%%%%%%%%%%%%%%%%%%%%%%%%%%%%%%%%%%%%%%%%%%%%%%%%%%%%%%%%%%%%%%%%%%%%%%%%%%%%%%%%%%%%%%%%%%%%%%%%%%%%%%%%%%%%%%%%%%%%%%%%%%%%%%%%%%%%%%%%%%%%%%%%%%%%%%%%%%%%%%%%%%%%%%%

\section{METHOD}
\label{sec_method}
\subsection{Robotic platform}
SpaceBok \cite{Spacebok} is a quadruped robot developed as a research platform to investigate dynamic locomotion in the context of planetary exploration. Its mass of \SI{22}{kg} and dimensions are comparable to that of a medium-sized dog. The robot has two actuators per leg (knee flexion/extension, hip flexion/extension), allowing the foot to move in the sagittal plane. Additionally, the legs' configuration uses a closed kinematic chain, where both actuators are used in parallel. This configuration is advantageous for the explosive generation of force, such as during jumping, but proved difficult to simulate accurately.
The actuators consist of brushless DC motors (T-Motor U8 KV85) in combination with a custom planetary gear transmission (1:9.55), which enables a maximum output torque of \unit[39.5]{Nm}. SpaceBok is equipped with an Intel i7 computer (Intel NUC) that executes our control software.

In all of our tasks, the robot must re-orient itself using its feet as reaction masses. As such, the achievable performance is directly dependent on the mass of the feet. In our experiments, we use high-inertia feet weighing \unit[330]{g} each, which in total sums up to the weight of a single axis reaction wheel \cite{kolvenbach2019iros}. This presents a considerable reduction in weight and mechanical complexity compared to the three reaction wheels needed to control all rotation axes.

\subsection{Simulation}
The training process requires days of real-time data and produces many failed episodes. As such, training on the physical robot is impractical. Even though recent work has shown that it is possible to train a policy directly on the physical robot \cite{realLearning}, the common practice is to train the network in simulation and transfer the trained policy to the hardware \cite{RubiksCube, AnimalImitation, AnymalRL, tan2018simtoreal}.

Simulating a jumping legged robot requires a rigid-body simulator with realistic contact handling, which should be as fast as possible and ideally multiple times faster than real-time. Additionally, since we train our policies with the specific objective of transferring to the physical robot, simulation accuracy is crucial. These two conflicting requirements need to be appropriately balanced.
\subsubsection{Simulator selection}
\label{sec_simulator_selection}
We compare two simulators: Mujoco (Version 2.0) \cite{Mujoco} and RaiSim (Version 0.7) \cite{Raisim}.
RaiSim is better suited for simulating hard contacts between the feet and the ground and can be vastly faster\footnote{The same training algorithm can run up to 10x faster using RaiSim. However, the simulation throughput depends more on interfacing than the actual simulator. Our training algorithms use Python libraries, while both simulators are written in C++. RaiSim comes with the RaiSimGym library, which allows the entire environment to be written in C++ and requires much less data copying.}. However, we find that it cannot handle the complex non-linearities of our problem. The self-reorientation task heavily depends on the exact conservation of the full system's momentum, which requires a precise integration method. Mujoco allows selecting between Euler's and 4\textsuperscript{th} order Runge-Kutta's (RK4) methods, while Raisim only uses Euler's method. We see that even with small time steps, Euler's method quickly diverges, leading to unrealistic behavior. Qualitatively, the robot starts rotating in the wrong direction and keeps rotating once the feet stop moving even though the angular momentum should be precisely zero. On the other hand, we see the exact conservation of angular momentum with RK4 integration. We therefore use Mujoco throughout this work.
\subsubsection{Closed kinematic chain simulation}
\begin{figure}[!bt]
\centering
\hspace*{5mm}\includegraphics[width=0.9\linewidth, trim={50, 10mm, 350mm, 10mm}, clip]{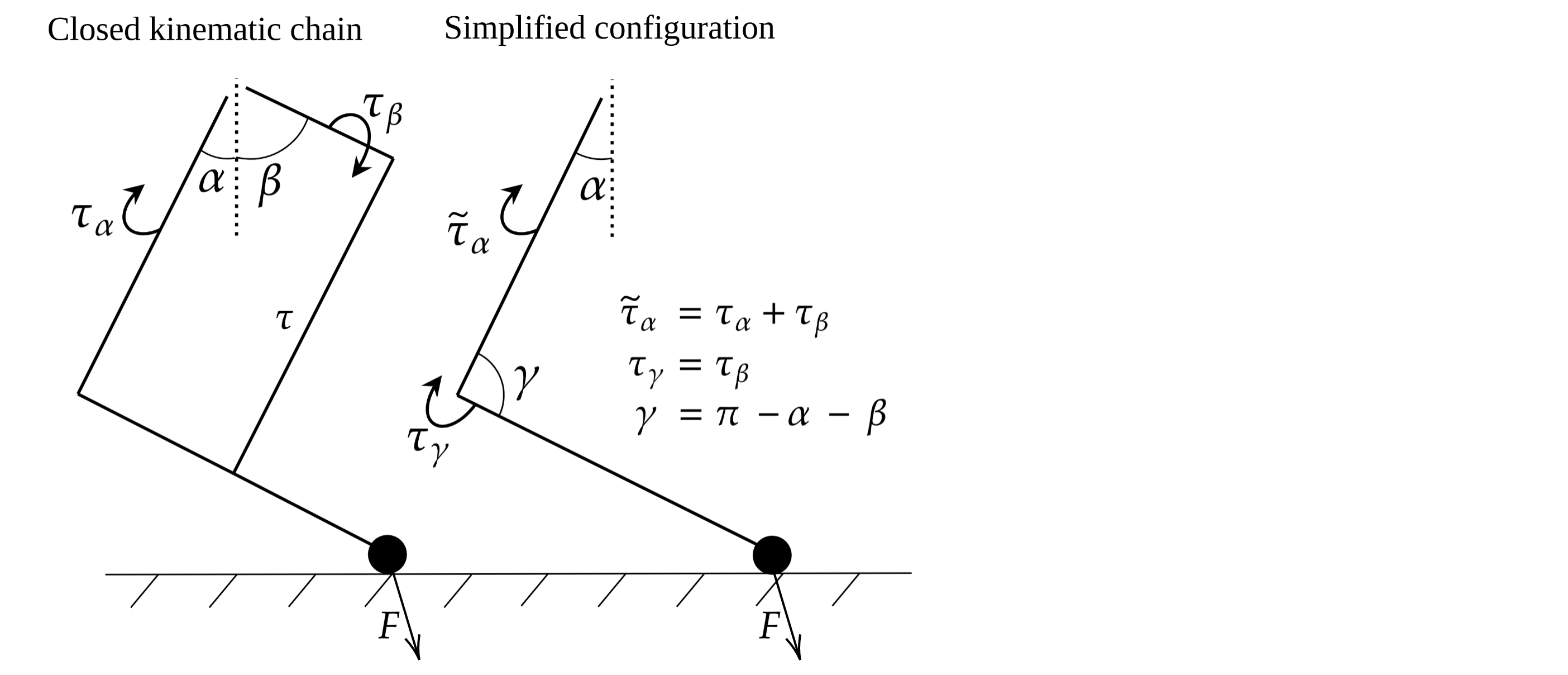}
\caption{Closed kinematic chain of the physical robot and the simplified configuration used for training.}
\label{fig_kinematic_chain}
\end{figure}
Unfortunately, standard rigid body simulators are meant to handle kinematic trees without closed chains. In Mujoco, simulating SpaceBok's parallel configuration requires additional constraints that cause simulation issues. With these constraints, even RK4 integration is not able to conserve angular momentum. RaiSim does not allow the simulation of closed kinematic chains.

Given these issues, we decide to train the robot using a simplified leg configuration with a single actuator in the shoulder and a second virtual actuator in the elbow, as shown in Figure \ref{fig_kinematic_chain}. In order to deploy trained policies on the physical robot, we compute the elbow joint angle $\gamma$ from the parallel configuration measurements $\alpha$, $\beta$ using
\begin{equation} \gamma = \pi - \alpha - \beta
\end{equation}
Additionally, we transform the torques produced by the PD controller on the simplified configuration $\Tilde{\tau}_{\alpha}$ and $\tau_{\gamma}$ to torques exerted on the parallel configuration $\tau_{\alpha}$ and $\tau_{\beta}$, producing the same forces on the feet and thus generating the same motion of the robot. In both configurations, forces applied on the feet can be calculated as
\begin{equation}
\mathbf{F} = \mathbf{J}^\top \boldsymbol{\tau}
\end{equation}
where $\mathbf{J}^\top$ are the transposed Jacobians $\mathbf{J}^\top_{s}$ or $\mathbf{J}^\top_{p}$, and $\boldsymbol{\tau}$ the respective joint torques $ \boldsymbol{\tau}_{s}=[\Tilde{\tau}_{\alpha}, \tau_{\gamma}]^\top$ and $\boldsymbol{\tau}_{p}=[\tau_
{\alpha}, \tau_{\beta}]^\top$.
Equating the forces in both case, we have
\begin{equation}
\boldsymbol{\tau}_{p} = \text{inv}(\mathbf{J}^{\top}_{p})\mathbf{J}^\top_{s}\boldsymbol{\tau}_{s}
\end{equation}
Computing $\text{inv}(\mathbf{J}^{\top}_{p})\mathbf{J}^\top_{s}$, we get a simple solution independent of joint angles:
\begin{equation}
\boldsymbol{\tau}_{p} = \begin{bmatrix}
1 & -1\\
0 & 1
\end{bmatrix}
\boldsymbol{\tau}_{s}
\end{equation}
\subsubsection{Actuator model}
\begin{figure}[!tb]
    \centering
    \includegraphics[width=0.65\linewidth]{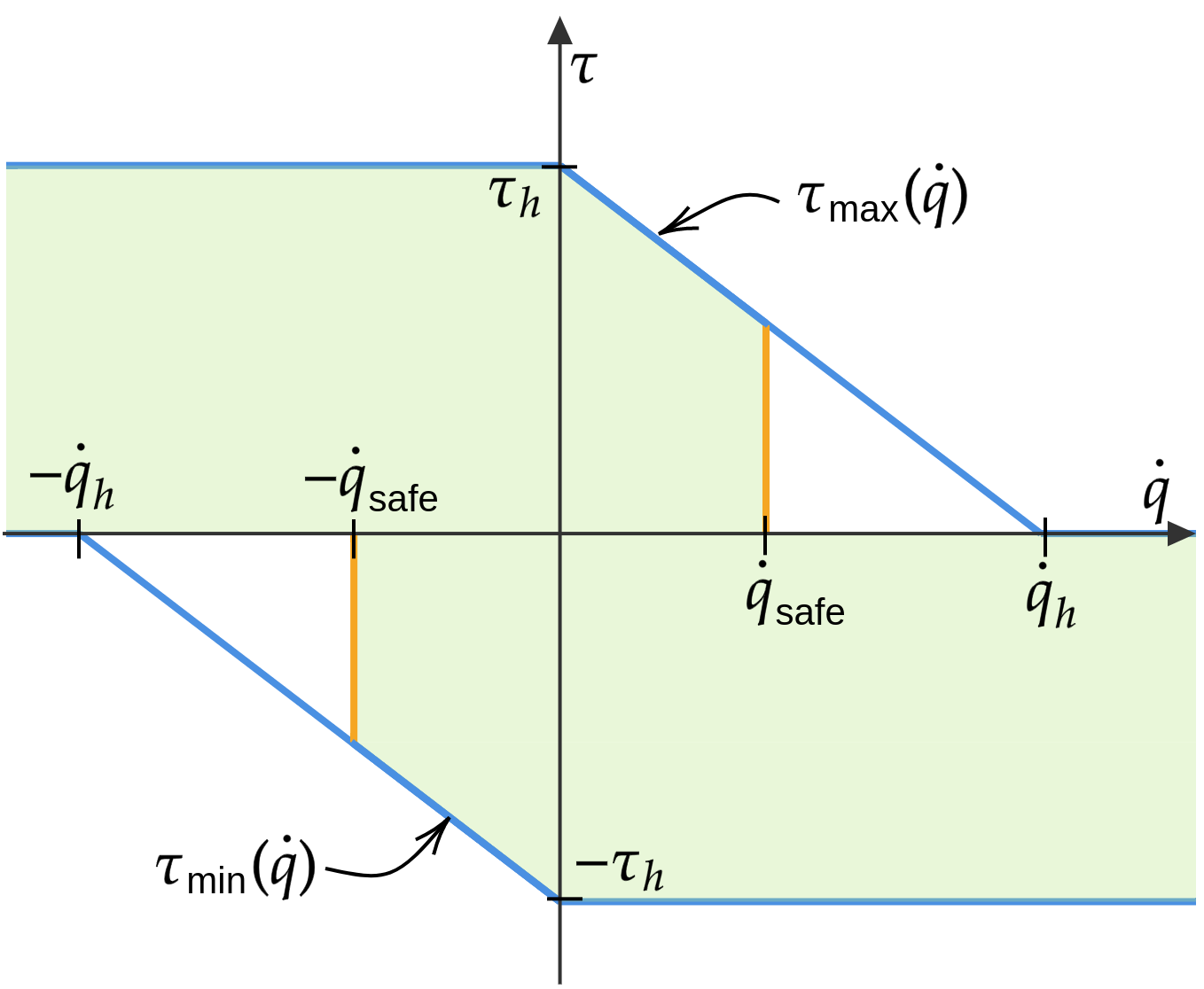}
    \caption{Allowed actuator torques (green region) as a function of joint velocity. $\tau_h$ and $\dot{q}_{h}$ are highest physically achievable torque and velocity of the actuator.}
    \label{fig_actuator_model}
\end{figure}
\label{sub_sub_sec_actuator_model}
We include a simple actuator model in our simulation. The maximum and minimum torques $\tau_{max}, \tau_{min}$ of a DC motor are affine functions of its current velocity. We model this by clipping the actuator torques to the $[\tau_{min}: \tau_{max}]$ range. Additionally, we want to limit joint velocities to have a safer transfer to the physical robot. \\
We achieve this by not allowing motors to further accelerate joints with velocities above a threshold $\dot{q}_{\textit{safe}}$. This corresponds to setting $\tau_{max}$ or $\tau_{min}$ to zero. Before being sent to the motors, torques computed by the PD controller are clipped to the closest point in the allowed region represented in Figure \ref{fig_actuator_model}.

The action space is defined as desired joint positions $\mathbf{q}_j^*$. The desired joint positions are transformed into actuator torques $\boldsymbol{\tau}$ by a joint space proportional-derivative (PD) controller:
\begin{equation}
\boldsymbol{\tau}^*=k_p(\mathbf{q}_j^*-\mathbf{q}_j) - k_d\mathbf{\dot{q}}_j
\end{equation}
This provides stability during training and allows us to adjust PD gains between simulation and deployment to achieve better tracking without retraining the policy.

\subsubsection{Self-collisions}
In order to solve our tasks, the robot must turn its feet quite aggressively, creating a serious risk of collisions between the legs. During training, collisions are detected by the simulator and lead to an episode termination with a large negative reward. To increase the safety margin, we add virtual spheres around each foot that can collide with the legs and each other, but not the base or the ground. This forces the feet to stay further apart, leading to a safer motion once deployed on the physical robot.

\subsection{Training}
\subsubsection{Reinforcement Learning}
We consider the standard reinforcement learning problem where an agent interacts with an environment. At each time step \textit{k}, the agent applies an action $\mathbf{a}_k$ based on which the environment updates its internal state $\mathbf{s}_k$, produces an observation $\mathbf{o}_k$, and provides a reward $r_k$. The agent's objective is to maximize the discounted sum of rewards $\sum_{k=0}^\infty \gamma^k r_k$ received from the environment, where $\gamma<1$ is a fixed parameter describing the relative importance of near and long-term rewards. The agent uses its internal policy to select the actions based on the observations. The policy represents a probability distribution parametrized by a neural network. In particular, in our case, the policy is a Gaussian distribution with diagonal covariance. The neural network produces both the mean $\boldsymbol{\mu}_\theta$ and the covariance $\boldsymbol{\sigma}_\theta$. While the mean depends on the observation, the covariance is an independent learned variable. Finally, we obtain
$\mathbf{a}_k \sim \mathcal{N}(\boldsymbol{\mu}_\theta(\mathbf{o}_k), \boldsymbol{\sigma}^2_\theta)$.
The training process optimizes the weights $\theta$ of this network to maximize the expected return.
\subsubsection{Policy optimization}
The policy's architecture is a neural network with three hidden layers of 128 neurons, resulting in approximately 35k parameters. The weights are learned using an off the shelf implementation \cite{stable-baselines} of the Proximal Policy Optimization (PPO) algorithm \cite{PPO}. We train all policies for 150 million steps, corresponding to 150000 episodes, 1500 policy updates, and 6 hours of training on a modern desktop computer\footnote{12 core AMD Ryzen 3900X, GTX 1070Ti, 32GB RAM, 96 parallel environments.}. Without domain randomization needed for sim-to-real transfer, policies can be trained up to three times faster. The training length is also increased for fine-tuning of the policy. After 100M steps, the learning rate is decreased via an exponential decay, providing a slight improvement in the final performance.
The set of tuned hyper-parameters can be found in Table \ref{table_hyper_param}.

\subsubsection{Sim-to-Real transfer}
\label{subsec_domain_rand}
Simulated environments are only a coarse representation of reality.
It is nearly impossible to simulate all of the complex dynamics of an electro-mechanical system. Furthermore, parameters such as masses, moment-of-inertia, and frictions coefficients are only approximations of the actual values. Since complex dynamics, not captured by simulators, mostly arise in the high-frequency regime, we push the policy towards smooth actions and enforce limits on the actuator velocities and torques.
We handle model mismatches between the simulation and the real robot using domain randomization \cite{DomainRandomization}. The policy is trained on many environments in parallel, each having a randomized set of simulation parameters. We randomize the weights of the feet ($\SI{330}{g} \pm30\%$), the intrinsic static friction of actuators ($\SI{0.45}{N/m} \pm50\%$) and the gains of the PD controller ($k_p = 20\pm20\%$, $k_d = 1\pm20\%$).
\subsubsection{Deployment}
Once fully trained, the network's weights are transferred to the robot without retraining or domain adaptation.
Target joint positions are constrained to respect the closed kinematic chain configuration, and desired torques can not exceed our actuator model's limits.
Additionally, observations are clipped to prevent unexpected behavior due to corrupted measurements. These limits can be adjusted without any retraining.
\begin{table}[tb!]\centering
\resizebox{\linewidth}{!}{
\begin{tabular}{rlrl} \\ \toprule
    Environment & & PPO &\\
    \midrule
        Steps per episode & 1K & Total Steps &  200M\\
        Simulation frequency & 400 Hz & Steps per update & 100K\\
        Policy frequency & 100 Hz & Batch size & 1K\\
        PD gains & P=20, D=1 & Discount factor & 0.9995\\
        Max joint torque $\tau_h$ & 25 Nm & Max grad norm & 10\\
        Joint velocity limit $\dot{q}_{\textit{safe}}$ & 10 rad/s & Entropy coef. & 0\\
 \bottomrule
\end{tabular}
}
\caption{Hyperparameters used for all tasks. All other PPO parameters are kept at the default values of the stable baselines implementation.}
\label{table_hyper_param}
% \vspace{-5mm}
\end{table}
%
%%%%%%%%%%%%%%%%%%%%%%%%%%%%%%%%%%%%%%%%%%%%%%%%%%%%%%%%%%%%%%%%%%%%%%%%%%%%%%%%%%%%%%%%%%%%%%%%%%%%%%%%%%%%%%%%%%%%%%%%%%%%%%%%%%%%%%%%%%%%%%%%%%%%%%%%%%%%%%%%%%%%%%%%%%%%%%%%%%%%%%%%%%%%%%%%%%%%%%
\subsection{Experimental Set-Up}
\label{subsec_test_bed}
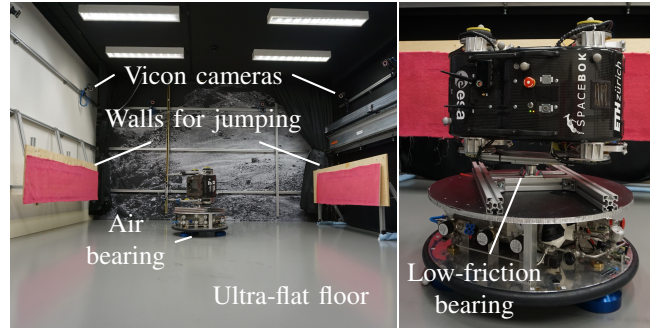
\begin{figure}[!t]
  \centering
    \usetikzlibrary{quotes,angles}
\begin{tikzpicture}
    \node[anchor=south west,inner sep=0] (image) at (0,0) {\includegraphics[width=0.58\linewidth, trim={1.5cm, 0, 1.5cm, 0},clip]{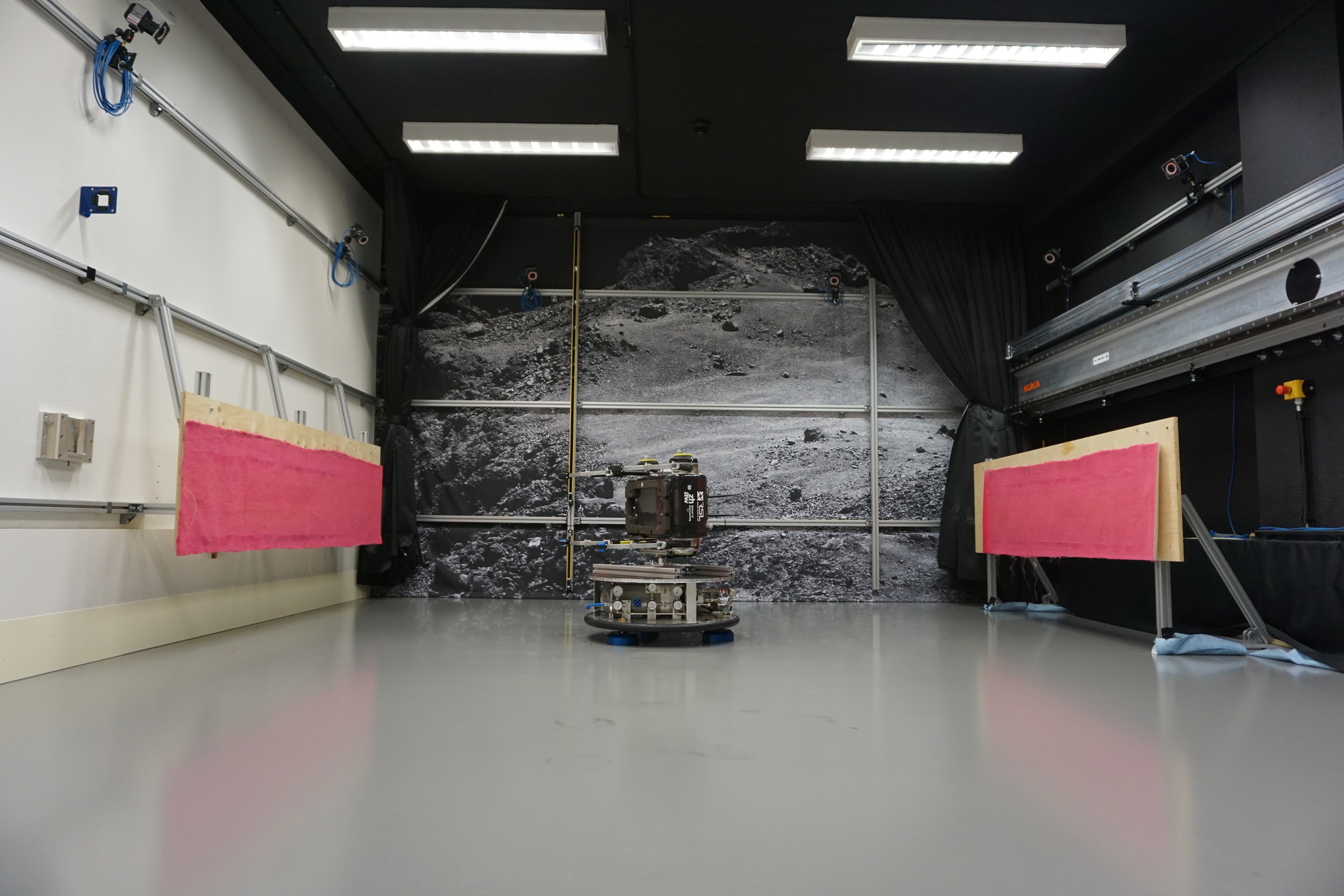}};
    \begin{scope}[x={(image.south east)},y={(image.north west)}]
        \node[anchor=south west, inner sep=0] (image2) at (1.006,0.0) {\includegraphics[width=0.375\linewidth, trim={2cm, 1.5cm, 1cm, 4.28cm}, clip]{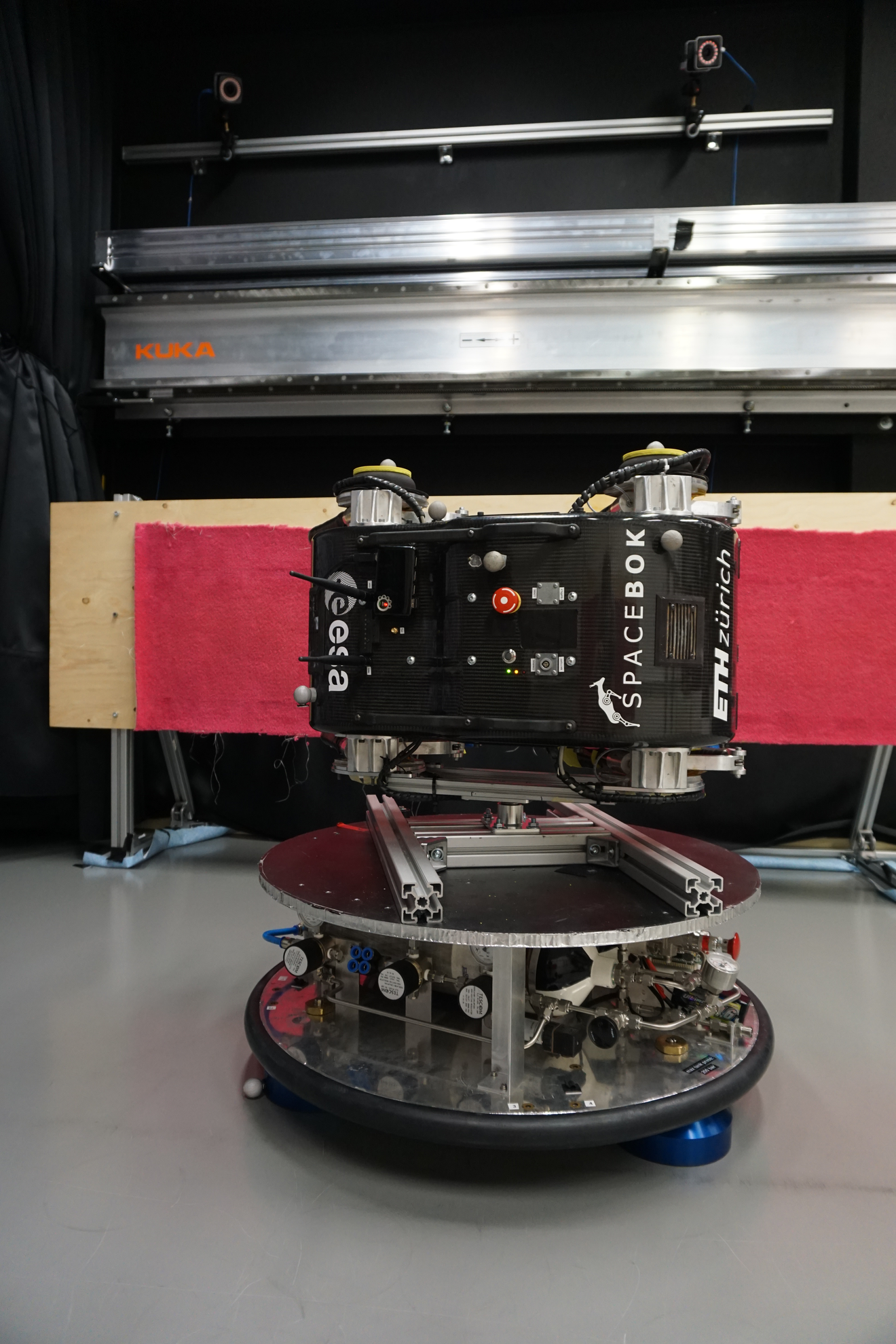}};
        \node [anchor=south, text=white,align=center] (bearing) at (1.2, 0.) {Low-friction\\ bearing};
        \node [anchor=south, text=white,align=center] (air-bearing) at (0.3, 0.15) {Air\\ bearing};
        \node [anchor=west, text=white,align=center] (floor) at (0.5,0.1) {Ultra-flat floor};
        \node [anchor=south, text=white, align=center] (walls) at (0.5, 0.57) {Walls for jumping};
        \node [anchor=south, text=white,align=center] (vicon) at (0.5, 0.72) {Vicon cameras};
        \draw [ thick, white] (bearing) -- (1.32,0.46);
        \draw [ thick, white] (air-bearing.east) -- (0.47,0.28);
        \draw [ thick, white] (walls) -- (0.22,0.5);
        \draw [ thick, white] (walls) -- (0.8,0.5);
        \draw [ thick, white] (vicon.west) -- (0.22,0.74);
        \draw [ thick, white] (vicon.east) -- (0.84,0.72);
    \end{scope}
\end{tikzpicture}
  \caption{SpaceBok on the experimental test-bed at the European Space Agency.} 
  \label{fig:esa set up}
\end{figure}
Simulating micro-gravity on Earth is not trivial. Our experimental approach consists in reducing the task to two dimensions in the robot's sagittal plane. Gravity offloading occurs by flipping the robot by \SI{90}{\degree} around its longitudinal axis. The lack of abduction in SpaceBok's legs is well suited for such experiments since it guarantees that gravity out of the sagittal plane does not affect the robot.

The European Space Technology and Research Center (ESTEC) has a unique facility developed for two-dimensional micro-gravity tests called ORBIT \cite{kolvenbach2016orbit}. Here, air-bearing platforms float friction-less above an ultra-flat floor. We fix the robot on top of one of those platforms via a rotational bearing, located in line with the principle axis of pitch-rotation of the robot (Figure \ref{fig:esa set up}). This allows the robot to move freely in its sagittal plane (via the air bearing platform) and turn around its pitch axis. The ball bearing allows us to decouple the rotational inertia of the air-bearing platform from the robot, but for jumps, we still need to account for the complete mass of the stack. In simulation, we add the platform's mass to the robot's base mass and leave the rotational inertia unchanged. Finally, an absolute motion capture system consisting of twelve Vicon cameras allows us to track the robot. The system is used to compute the position, orientation, and velocities of the robot's base and provide these measurements to the policy.

One pitfall of the described set-up is that the robot's ball bearing attachment is below the center of mass, which creates large bending moments during aggressive feet motion. Due to the attachment mechanism's insufficient stiffness, we observe significant shaking of the robot's base. We overcome this limitation by having a similar set-up in simulation, with the base attached via a spring to a platform free-floating in the sagittal plane, and training the policy to reduce the shaking. In this configuration, the position of the attachment point between the platform and the robot highly influences the system's dynamics. On the physical robot, we aim to place that point under the center of mass and train the policy to be robust to small offsets by adding the attachment point to the set of randomized variable described in Section \ref{subsec_domain_rand}, with a range of $\pm\SI{0.1}{m}$ around the center of mass.

%%%%%%%%%%%%%%%%%%%%%%%%%%%%%%%%%%%%%%%%%%%%%%%%%%%%%%%%%%%%%%%%%%%%%%%%%%%%%%%%%%%%%%%%%%%%%%%%%%%%%%%%%%%%%%%%%%%%%%%%%%%%%%%%%%%%%%%%%%%%%%%%%%%%%%%%%%%%%%%%%%%%%%%%%%%%%%%%%%%%%%%%%%%%%%%%%%%%%%%%
\begin{table}[b!]\centering
\begin{tabular}{rlrl}
    \toprule
     $r_{\textit{orient,2D}}$ & $-|\phi|$ & $r_{\textit{action clip}}$  & $-||\mathbf{q}_j^* - \mathbf{q}_{j,\textit{clipped}}||^2$\\
     $r_{\textit{orient,3D}}$ & $-|\textit{angle}(\mathbf{q}_b)|$ & $r_{\textit{torque clip}}$ & $-||\boldsymbol{\tau}^* - \boldsymbol{\tau}_{\textit{clipped}}||^2$\\
     $r_{\textit{jump}}$ & $\exp(-\frac{||\mathbf{v}^*_b - \mathbf{v}_b||^2}{0.2^2})$ & $r_{\textit{base acc}}$ & $-||\dot{\mathbf{v}}_b||^2$  \\
     $r_{\textit{torque}}$ & $-||\boldsymbol{\tau}^*||^2$\\
 \bottomrule
\end{tabular}
\caption{Definition of reward terms.}
\label{table_rewards}
\vspace{-5mm}
\end{table}
\begin{table*}[!ht]\centering
\resizebox{\linewidth}{!}{
\begin{tabular}{cccccccc} \\ \toprule
     & $\mathbf{r}_b \, [m]$ & $\mathbf{v}_b \, [m/s]$ & roll, pitch, yaw $[\textit{rad}]$ & $\boldsymbol{\omega}_b \, [\textit{rad}/s]$ & $\mathbf{v}^*_b \, [m/s]$ & $q_j \,[\textit{rad}]$ & $\dot{q}_j \, [\textit{rad}/s]$ \\ \midrule
    2D orientation & $(0,0,2)$ & $\mathbf{0}$ & $\eta(\text{-},[-\pi: \pi],\text{-})$ & $\mathbf{0}$ & - & IK & $\mathbf{0}$ \\
    2D jumping & $(0,0,3)$ & $(\eta[-0.5:0.5],\text{-},[-1:0])$ & $\eta(\text{-},[-\pi: \pi],\text{-})$ & $\eta(\text{-},[-0.25:0.25],\text{-})$ & $\eta([-0.5:0.5],\text{-},[0.25:1])$ & IK & $\mathbf{0}$ \\
    3D orientation & $(0,0,2)$ & $\mathbf{0}$ & $\eta([-\pi: \pi]\times3)$ & $\mathbf{0}$ & - & IK & $\mathbf{0}$ \\
    3D landing & $(0,0,3)$ & $(0,0,-1)$ & $\eta([-\pi: \pi]\times3)$ & $\mathbf{0}$ & $\mathbf{0}$ & IK & $\mathbf{0}$ \\
 \bottomrule
\end{tabular}
}
\caption{Initial state distributions for all tasks. For ranges, the value is sampled uniformly.  $\eta$ is the curriculum parameter, starting at $0.25$ (3D tasks) or $0.5$ (2D tasks) and slowly increasing to 1 as training progresses. We use Inverse Kinematics (IK) to initialize the joints in a configuration that places the feet inside a circle of \SI{0.1}{m} radius around the nominal position.}
\label{table_init_states}
\end{table*}

\section{2D Attitude control}
\label{sec_2d_attitude_control}
The first task focuses purely on controlling the pitch axis without any ground contact, mimicking a free-floating scenario in space. During this task, the robot transitions from a random initial orientation to a target orientation as quickly as possible.

Due to the symmetry of the problem, we do not need to train the policy on all possible target angles but rather to always reach a zero angle from randomized initial conditions. Once deployed, we use the desired target to offset the angle in the observation. The policy sees this angle as an error and will reduce it to zero by rotating the robot to the desired orientation. 

Reward terms are defined in Table \ref{table_rewards} and initial states distributions are presented in Table \ref{table_init_states}.
\subsubsection{Observation space}
The observations given to the policy include 
     joint positions $\mathbf{q}_j$,
     joint velocities $\dot{\mathbf{q}_j}$,
     base pitch angle $\phi_b$,
     base pitch angular velocity $\dot{\phi}_b$.
All positions and velocities are given as measured at the last time step, no history is provided.
\subsubsection{Action space}
The output of the policy is interpreted as desired joint positions, $\mathbf{q}_j^*$. Desired joint positions and corresponding torques are processed to ensure that limits of the closed kinematic chain and our actuator model are respected.

We develop two levels of complexity for this task. First, the policy only sees and controls the robot's left legs while the right side follows the same trajectory. This simplifies the training process but limits the performance of the policy. We then train a policy with individual control over each leg and compare each case to simple handcrafted trajectories.
\subsubsection{Reward}
The total reward consist of a weighted sum of four elements with $(c_0, c_1, c_2, c_3, c_4) = (\frac{1}{200}, 1, \frac{1}{100}, \frac{1}{3}, \frac{1}{3})$:
\begin{equation}\label{rew_weighted_sum}
r = c_0(c_1 r_{\textit{orient,2D}} + c_2 r_{\textit{base acc}} + c_3 r_{\textit{action clip}} + c_4 r_{\textit{torque clip}})
\end{equation}

The main reward $r_{\textit{orient,2D}}$ pushes the policy to reduce the pitch angle to zero. The second term $r_{\textit{base acc}}$ penalizes base accelerations, therefore encouraging the policy to minimize the shaking of the robot.

As described above, the geometry of the robot's leg imposes constraints on the possible joint positions, but the policy can select positions outside of these constraints. When this happens, the erroneous actions are clipped to the closest allowed configuration, and the $r_{\textit{action clip}}$ penalty is added to the reward.
Finally, the torques sent to the actuators are limited to represent the limits of the physical motors and to prevent high joint velocities, as described in section \ref{sub_sub_sec_actuator_model}. Similarly to action clipping, when the PD controller requires torques that lie outside the limits, they are clipped to the closest allowed values, and the $r_{\textit{torque clip}}$ penalty is added to the reward.

\subsection{Training}
Achieving successful training requires extensive hyper-parameter tuning, reward shaping, and the use of a curriculum. Before proper tuning, we obtain multiple policies that are not moving at all when the initial angle is too large. They prefer accumulating the orientation error penalty rather than risking a collision. More successful policies learn to reorient the robot in only one direction, either clockwise or counterclockwise. Finally, policies turning in both directions tend to stop within \SI{10}{\degree} of the target and do not correct the final small error. We solve these problems using the following techniques:
\begin{enumerate}
    \item We use a curriculum of initial states. At the beginning of training, the robot starts with simple initial states. Once the policy learns to solve the simpler case, the complexity is slowly increased.
    \item We increase the number of steps per update of the policy together with the batch size to $100000$ and $1000$ steps, respectively. This leads to slower but more stable training, allowing the policy to find complex solutions.
    \item We use the absolute value of the orientation error $|\phi|$ as penalty since it is better at pushing small values towards zero than the squared error $\phi^2$. We also use a large discount factor of $\gamma=0.9995$ in order to penalize policies that stop moving without reaching the target.
\end{enumerate}

\subsection{Simulation performance}
\begin{figure}[!tb]
\centering
    \subfloat{\includegraphics[width=\linewidth]{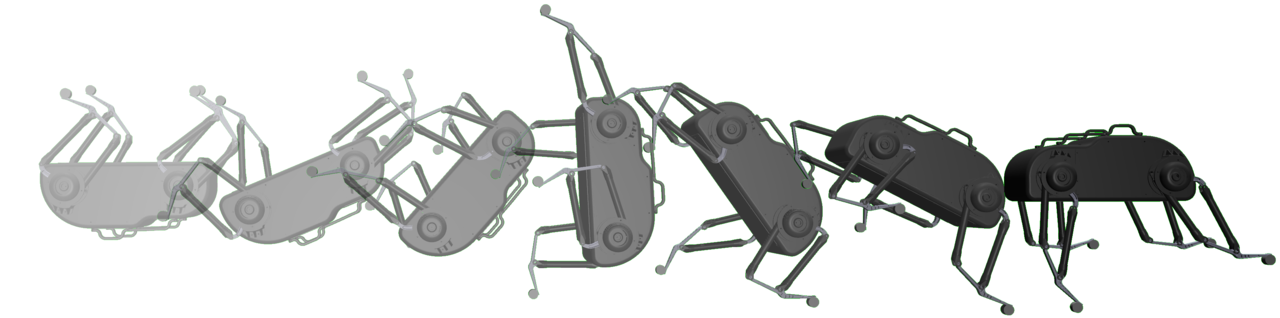}}
  \caption{Example of a reorientation episode. The asymmetric learned policy performs a \SI{180}{\degree} flip in approximately 4s and stops within \SI{0.5}{\degree} of the target.}
  \label{fig_2d_orient_picture}
\end{figure}
\begin{figure}[!tb]
  \centering
    \includegraphics[width=\linewidth, trim={5mm, 0, 5mm, 0},clip]{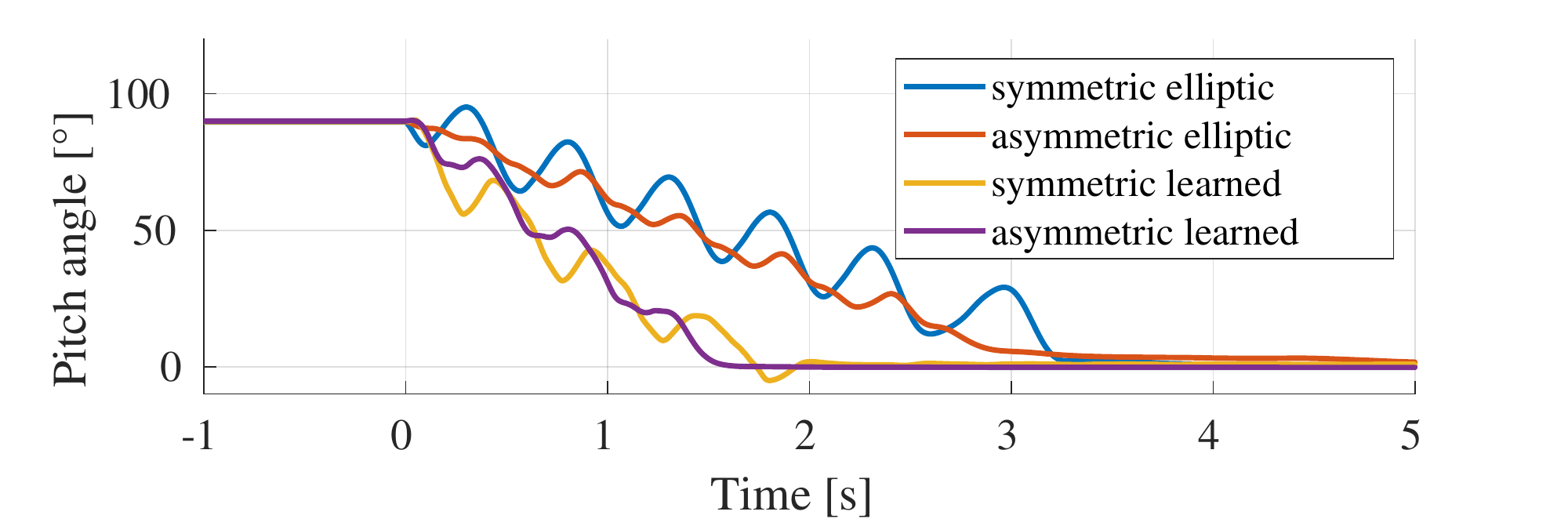}
  \caption{Comparison of two baselines with two versions of learned policies. Both policies highly outperform the baseline, while asymmetric motion reduces back-and-forth motion.} 
  \label{fig_2d_orient_plot_baselines}
  \vspace{-5mm}
\end{figure}

The policy trained on 2D orientation is evaluated by assessing the tracking performance of the target pitch angle. The target is modified in steps of \SI{90}{\degree} and \SI{180}{\degree}. Figure \ref{fig_2d_orient_picture} shows the robot performing a \SI{180}{\degree} flip. As a baseline, we use a handcrafted policy turning the feet in an elliptic trajectory with a proportional controller providing the frequency of that motion. Similarly to the two versions of the learning task, we use two baseline versions. One with identical motion between left and right legs and one with an angular offset between each leg. Figure \ref{fig_2d_orient_plot_baselines} demonstrates that both learned policies highly outperform our baseline motion. The asymmetric learned policy manages to rotate the robot by \SI{90}{\degree} in less than \SI{1.5}{s} while the baselines take more than \SI{3}{s}.  In both cases, asymmetric motions produce fewer back-and-forth movements on the robot, resulting in a smoother trajectory.

\subsubsection{Effect of feet weights}
\begin{figure}[bt!]
  \centering
    \includegraphics[width=\linewidth, trim={5mm, 0, 5mm, 0},clip]{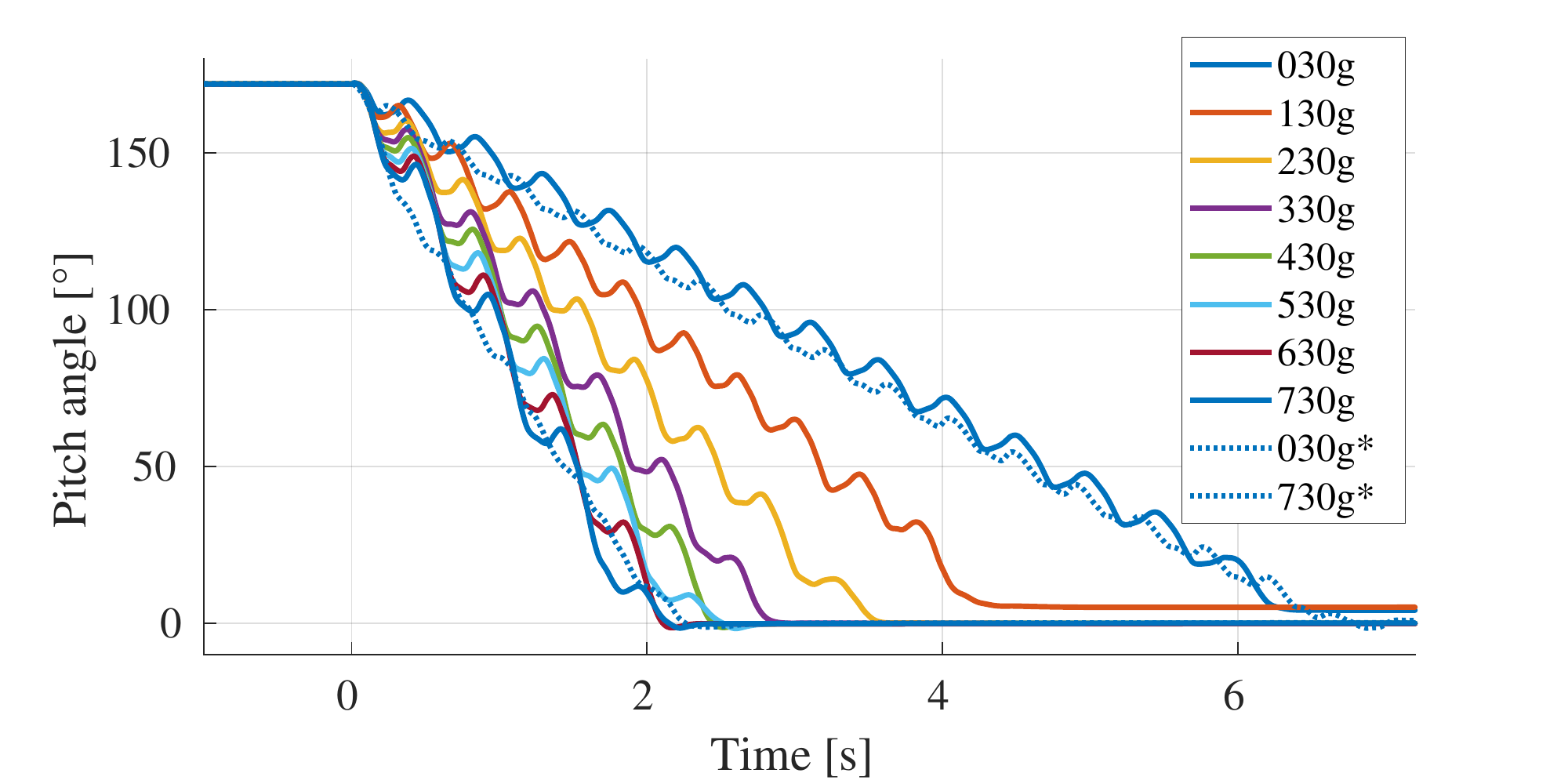}
  \caption{Evolution of the pitch angle during a 2D re-orientation episode with different weights added to the feet. The policy is trained on weights between \SI{230}{g} and \SI{430}{g}, but generalizes well beyond this range. Curves marked with * represent a policy that was trained with the correct weight of \SI{30}{g} and \SI{730}{g} respectively.} 
  \label{fig_weights}
\end{figure}

We explore the effect of the high-inertia feet. During training, the added weight is randomized, as described in Section \ref{subsec_domain_rand}. Once the policy is trained, we analyze its generalization capabilities by changing the weight even further. Figure \ref{fig_weights} shows that the policy performs well with weight from \SI{30}{g} up to \SI{730}{g} without any retraining. Interestingly, The performance is also very close to policies trained with a single correct weight, showing that the learned behavior scales well across these changing dynamics, and retraining is unnecessary.

Figure \ref{fig_weights} also allows us to verify the choice of weight added to the physical robot. The increase in re-orientation performance needs to be balanced with the detrimental increase in the total robot's weight. As expected, higher weights allow the policy to perform faster flips, but the performance plateaus around \SI{600}{g}. A weight between \SI{200}{g} and \SI{400}{g} seems to be a reasonable choice given the respective performances. 

\subsection{Deployment performance}
\begin{figure}[!tb]
  \centering
    \includegraphics[width=\linewidth, trim={4mm, 0, 5mm, 0},clip]{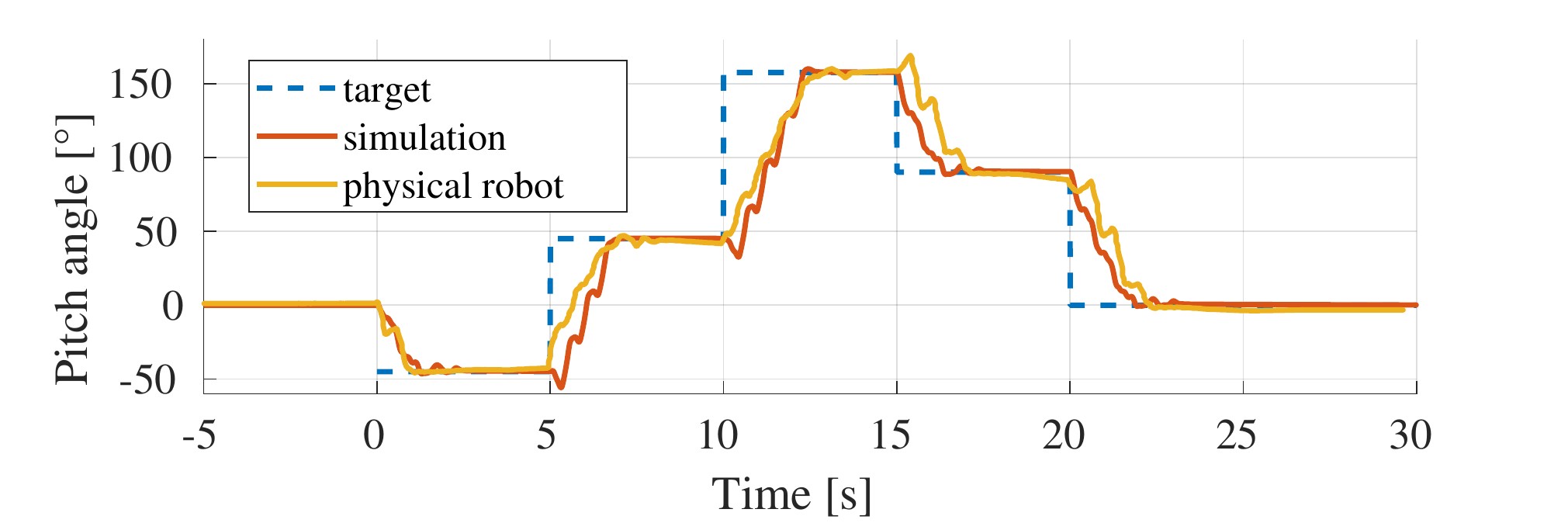}
  \caption{Asymmetric policy deployed in simulation and on the physical robot, tracking five changes in the pitch angle target over 30 seconds.} 
  \label{fig_2d_orient_steps}
\end{figure}

The asymmetric policy is deployed on the set-up described in section \ref{subsec_test_bed}. The air-bearing is turned off, leaving the robot's base with a single degree of freedom around its pitch axis. The policy is tasked to follow multiple changes in the pitch target. It manages to solve the task successfully despite the shaking introduced by the physical set-up. We evaluate the performance gap between simulation and reality by setting the same initial conditions and targets in simulation and comparing the simulated and experimental roll-outs in Figure \ref{fig_2d_orient_steps}. \\
We see a surprisingly small gap between simulation and reality. In both cases, the robot requires less than \SI{2.5}{s} to perform a 90° rotation and stabilize. Overall, the deployed policy quickly reaches any desired target and can also recover when the robot is manually pushed even though it was never trained in this scenario.
%%%%%%%%%%%%%%%%%%%%%%%%%%%%%%%%%%%%%%%%%%%%%%%%%%%%%%%%%%%%%%%%%%%%%%%%%%%%%%%%%%%%%%%%%%%%%%%%%%%%%%%%%%%%%%%%%%%%%%%%%%%%%%%%%%%%%%%%%%%%%%%%%%%%%%%%%%%%%%%%%%%%%%%%%%%%%%%%%%%%%%%%%%%%%%%%%%%%%%%%
\section{2D Jumping}\label{sec_2d_jump}
The second task combines jumping and reorientation. The robot starts a few meters away from the wall (as seen in Figure \ref{fig:esa set up}), with a perpendicular velocity of up to \SI{1}{m/s} and an angle up to \SI{180}{\degree}. It must reorient itself, land, and jump back towards a target direction. 
\subsubsection{Observation space}
We extend the observations of the first task by adding the height of the robot's base from the ground $b_z$, base linear velocity $\mathbf{v}_b$, and desired base velocity $\mathbf{v}^*_b$.
\subsubsection{Action space}
To keep the jumping phase physically realistic, we need symmetric forces between the left and right legs. We achieve this by constraining the policy similarly to the symmetric policy of the previous task. 
\subsubsection{Reward}
The reward includes all terms of the previous task and is also extended with two new elements with respective coefficients $c_5=1$ and $c_6=\frac{1}{30}$. The coefficient of base acceleration is increased to $c_2=\frac{1}{20}$ in order to emphasize the importance of smooth jumps. We have
\begin{equation}\label{rew_weighted_sum_jump}
\begin{split}
r = c_0(c_1 r_{\textit{orient,2D}} + c_2 r_{\textit{base acc}} + c_3 r_{\textit{action clip}}\\+ c_4 r_{\textit{torque clip}} + c_5 r_{\textit{jump}} + c_6 r_{\textit{torques}})
\end{split}
\end{equation}

The first new element encourages the policy to track the commanded jump direction and velocity. An exponential kernel is used to transform the tracking error into a positive reward. 

During the landing phase, we notice high-frequency oscillations in the actions. These oscillations do not help to solve the task, so we encourage the policy to produce smoother actions by penalizing actuator torques with $r_{\textit{torques}}$.

\subsection{Training}
For the 2D jumping task, we use all training techniques that proved useful for 2D orientation, and we apply a curriculum on initial and target base velocities  $\mathbf{v}_{b,0}$, $\mathbf{v}_b^*$.
\subsubsection{Simulation performance}
A policy trained on the jumping task can successfully re-orient the robot, land, and jump back in a given direction. It can land from complex states where it does not have time to completely re-orient itself and learns to correct the trajectory after an imperfect jump by quickly touching the ground a second time. 

We replicate our experimental set-up by adding another wall above the robot at a height $d_{w}=\SI{4}{m}$. We then select the wall that the robot is going to hit next and transform exteroceptive observations to show it as the ground to the policy. Specifically, when $v_{b,z}>0$ and $b_{z}>\SI{0.7}{m}$, the observation is transformed into $\Tilde{\mathbf{o}}$, which differs from the original observation $\mathbf{o}$ by
\begin{equation}
    [\Tilde{v}_{b,x}, \Tilde{v}_{b,z}, \Tilde{v}^*_x] = -[v_{b,x}, v_{b,z}, v^*_x]
    \end{equation}
    
    \begin{equation}
    \Tilde{b}_z = d_{w} - b_z
    \end{equation}
    
    \begin{equation}
    \Tilde{\phi} = \frac{\pi}{2} + \phi 
\label{eq:1}
\end{equation}

We obtain a policy jumping from wall to wall while performing flips in between. It performs 100 jumps in a row without failure.
\begin{figure}[!tb]
    \centering
    \subfloat[][]{{\includegraphics[width=0.45\linewidth]{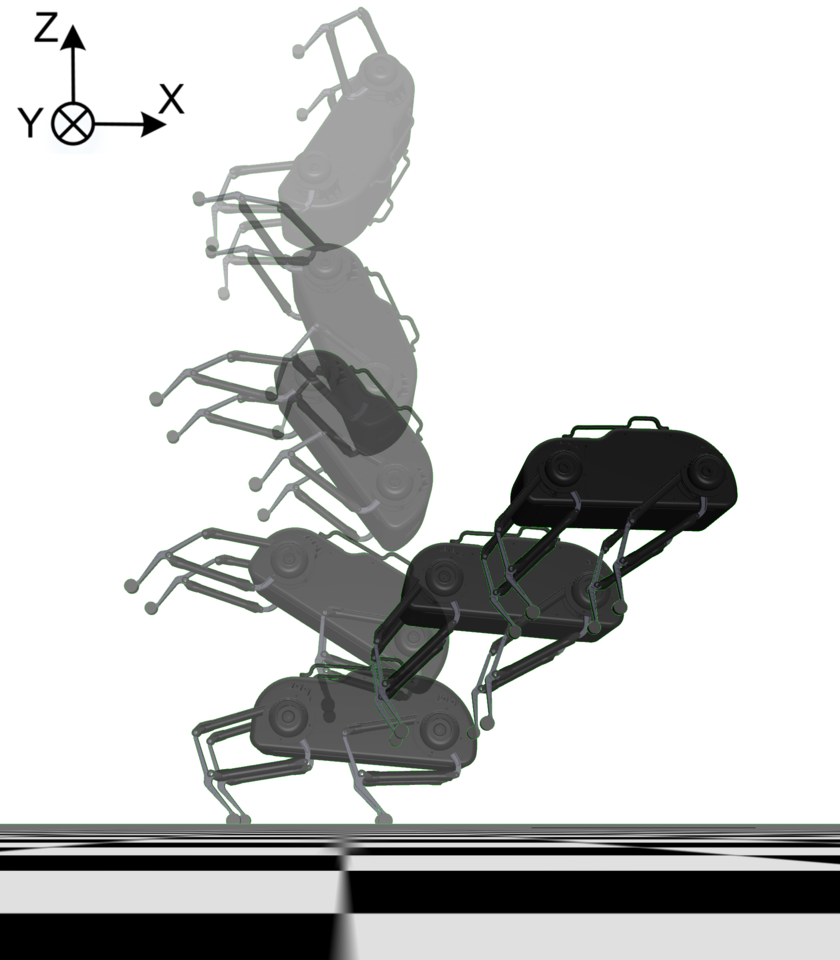}}}
    \,
  \subfloat[][]{{\includegraphics[width=0.45\linewidth]{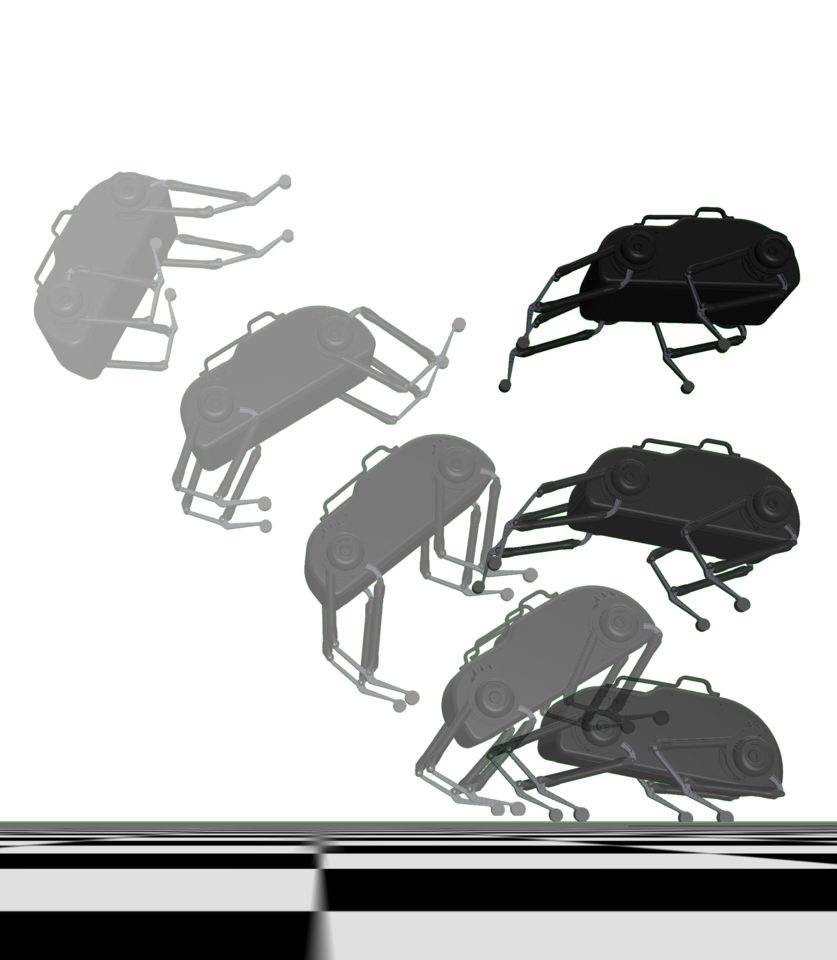}}}
    \caption{Two examples of a 2D orientation and jumping episodes. The robot starts with a velocity of 1m/s towards the ground. It reorients itself before landing and jumping with a target velocity vector of $\mathbf{v}^*_b=(0.5, 0, 0.5)[\textit{m/s}]$ for episode (a) and $\mathbf{v}^*_b=(0, 0, 1)[\textit{m/s}]$ for episode (b). }%
    \label{fig_2d_orientjumpvertical}%
    \vspace{-5mm}
\end{figure}
\subsection{Deployment performance}
\begin{figure}[!tb]
    \centering
    \subfloat[][]{{\includegraphics[width=0.45\linewidth, trim={0, 1.26cm, 0, 0cm},clip]{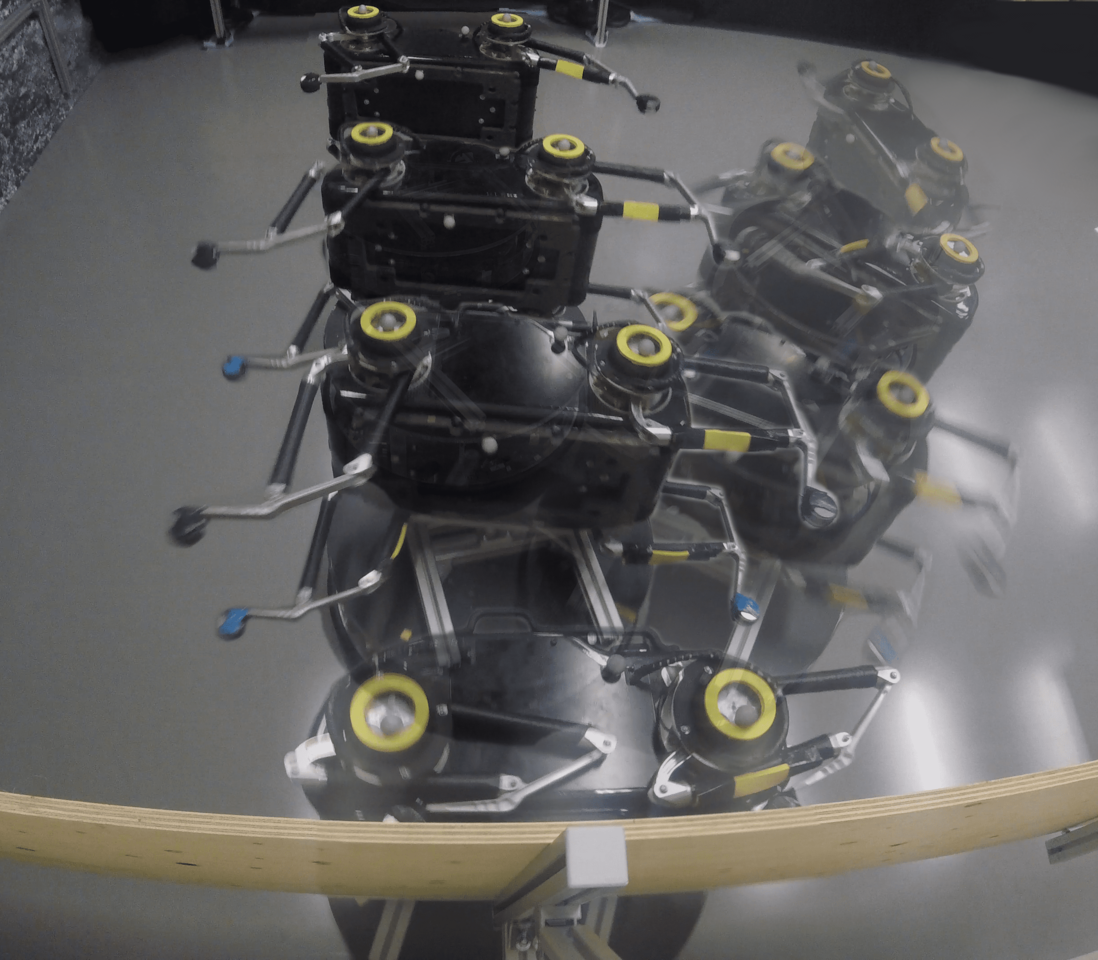}}}
    \,
  \subfloat[][]{{\includegraphics[width=0.45\linewidth, trim={1cm, 0 2cm, 0},clip]{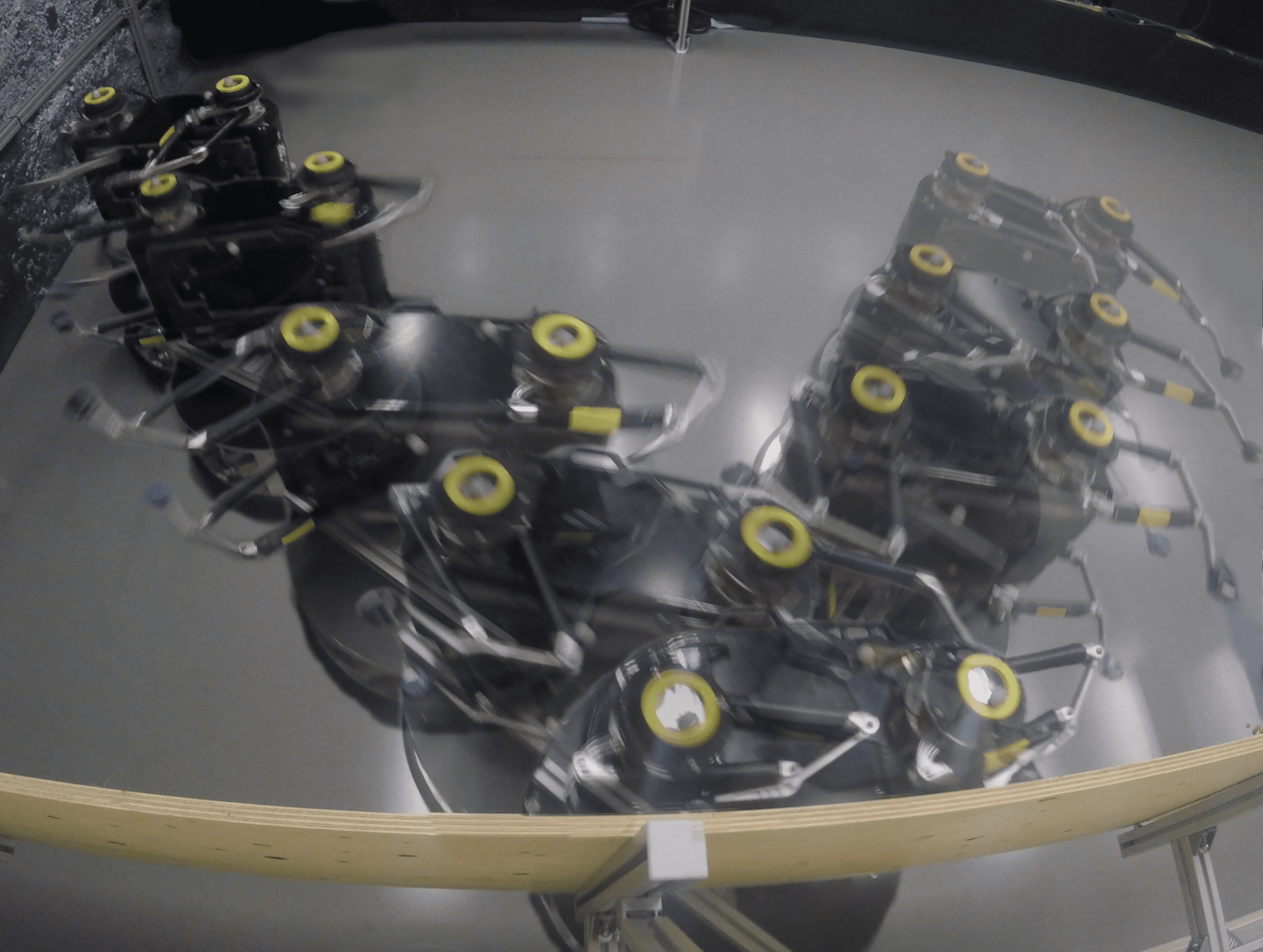}}}
    \caption{Jumping policy deployed on the physical robot. The robot is pushed manually towards the wall with $v_{b,x} \approx \SI{0.4}{m/s}$. The target jump velocities are $\mathbf{v}^*_b=(0, 0, 0.5)[\textit{m/s}]$ for episode (a) and $\mathbf{v}^*_b=(0.5, 0, 0.5)[\textit{m/s}]$ for episode (b). The policy manages to compensate the original momentum and jump in the correct direction. In these episodes the observations are not mirrored to the opposite wall.}%
    \label{fig:2d_orient_eval_real}%
\end{figure}

We notice that due to shaking described in Section \ref{subsec_test_bed}, even after filtering, the measurements of both linear and angular velocities on the testbed are very noisy. In particular, the noisy angular velocity $\dot{\phi}$ causes instabilities in the policy. Simply removing this measurement from the observation space, the policy learns to solve the task without $\dot{\phi}$. Once deployed, this policy is far superior in both stability and target tracking performance. Results of the simulation performance can be seen in Figure \ref{fig_2d_orientjumpvertical}. We deploy the trained policy on the physical robot. This time, the air-bearing is turned-on such that the robot floats with minimal friction. We begin our tests by manually throwing the robot towards the wall and catching it after the jump (Figure \ref{fig:2d_orient_eval_real}). \\
We assess the target velocity tracking performance by throwing the robot with similar initial conditions while varying the commanded targets. The simulation to reality gap is more noticeable in this experiment. Whereas in simulation, the jump velocity error is below $\pm \SI{8}{\percent}$, on the physical robot it increases to $\pm \SI{25}{\percent}$, most probably due to imperfect modeling of the friction forces between the feet and the walls. In further experiments, these results can be improved by randomizing the friction coefficients of the walls and tuning the contact model of the simulator, but also by redesigning the feet in order to obtain more consistent contacts on the real robot.

\begin{figure}[tb!]
  \centering
\includegraphics[width=0.9\linewidth]{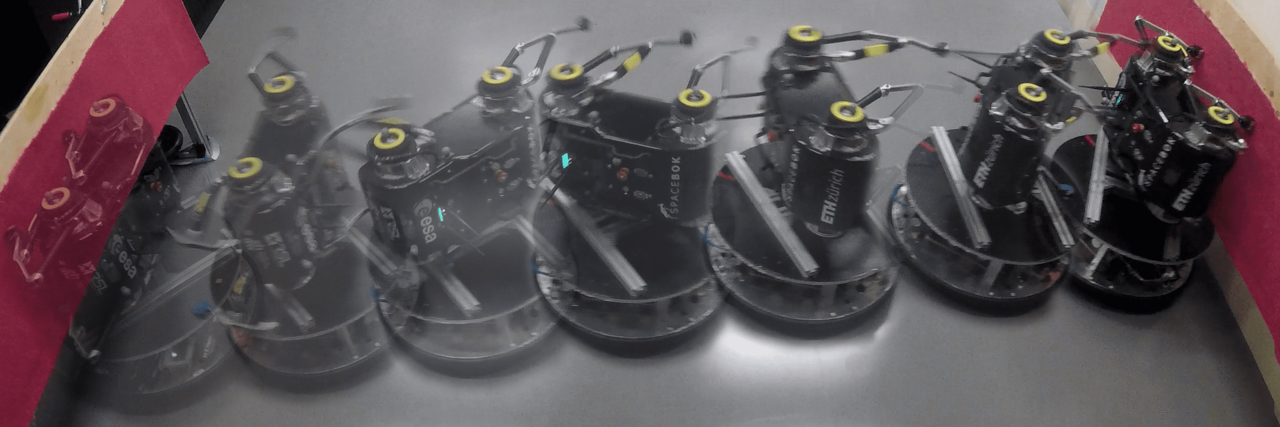}
  \caption{Jumping policy deployed on the physical robot with observation mirrored to the wall which the robot will hit next. The robot successfully jumps, flips during the flight and lands on the opposite side separated by \SI{4}{m}. It then jumps back in the other direction and the process repeats.} 
  \label{fig:2d_orient_eval_realwall}
  \vspace{-3mm}
\end{figure}

We further test the policy's robustness by turning on the mirroring of observations and letting the robot perform consecutive jumps between the walls (Figure \ref{fig:2d_orient_eval_realwall}). Due to the limited size of the jumping area, the robot needs to correct it's offset after each jump. We set $v^*_x$ such that $v^*$ points towards the center of the opposite wall, using
\begin{equation}
v^*_x=-\frac{b_{x}}{d_w}v^*_z
\end{equation}
The policy achieves dozens of consecutive jumps without failure. It selects the rotation direction based on the conditions after the jump and can even recover from situations with incomplete rotations using the landing wall to make last-second corrections.
%
%%%%%%%%%%%%%%%%%%%%%%%%%%%%%%%%%%%%%%%%%%%%%%%%%%%%%%%%%%%%%%%%%%%%%%%%%%%%%%%%%%%%%%%%%%%%%%%%%%%%%%%%%%%%%%%%%%%%%%%%%%%%%%%%%%%%%%%%%%%%%%%%%%%%%%%%%%%%%%%%%%%%%%%%%%%%%%%%%%%%%%%%%%%%%%%%%%%%%%

\section{3D Attitude control}\label{sec_3d_attitude_control}
We now increase the complexity and apply our approach to three-dimensional environments.

SpaceBok's lack of abduction highly reduces the control authority over the roll and yaw axis. We train a policy that controls all axis with the current leg configuration, but notice a big limitation in its performance, a \SI{90}{\deg} roll or yaw rotation taking over \SI{6}{s} (Figure \ref{fig_3dof_vs_2dof}). In order to obtain more practical results, we create a new simulated version of SpaceBok with extra actuators providing hip abduction/adduction.
\subsubsection{Observation space}
This task has a very similar observation space to its 2D counterpart. The only difference is that we now represent the base orientation with a quaternion $\mathbf{q}_b$, and the angular velocity is now a three-dimensional vector $\boldsymbol{\omega}_b$.
\subsubsection{Action space}
3D reorientation does not allow any symmetry between the left and right legs. In this case, the policy sees and controls all legs individually.
\subsubsection{Reward}
We use the same rewards as in Equation \ref{rew_weighted_sum}, but now obtain $r_{\textit{orient,3D}}$, by computing the angle-axis representation of the quaternion and penalizing the angle as before.

\subsection{Training}
In the 3D case, we find that the training is very unstable, and we need to apply a much slower curriculum starting with very small orientation errors. Even so, the reward oscillates and even goes below a \textit{do nothing} baseline. We see that these policies quickly learn to turn the feet to produce a base rotation but struggle to turn it in the correct direction, producing rewards worse than a policy that would not move at all. This is most probably related to the high coupling between rotations on all axis. After $\sim50\%$ of the training steps, the reward suddenly increases, and a solution that correctly reorients the robot is found. We can then slowly increase the range of initial states and train the policy to handle all possible orientations.
\subsection{Simulation performance}
\begin{figure}[!tb]
    \centering
    \subfloat{{\includegraphics[width=1.0\linewidth]{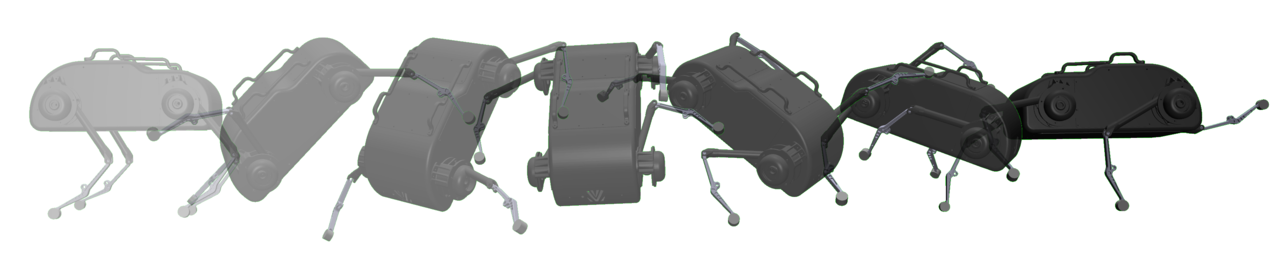}}}
    \vspace{-4mm}
    \subfloat{{\includegraphics[width=\linewidth, trim={5mm, 0, 5mm, 0},clip]{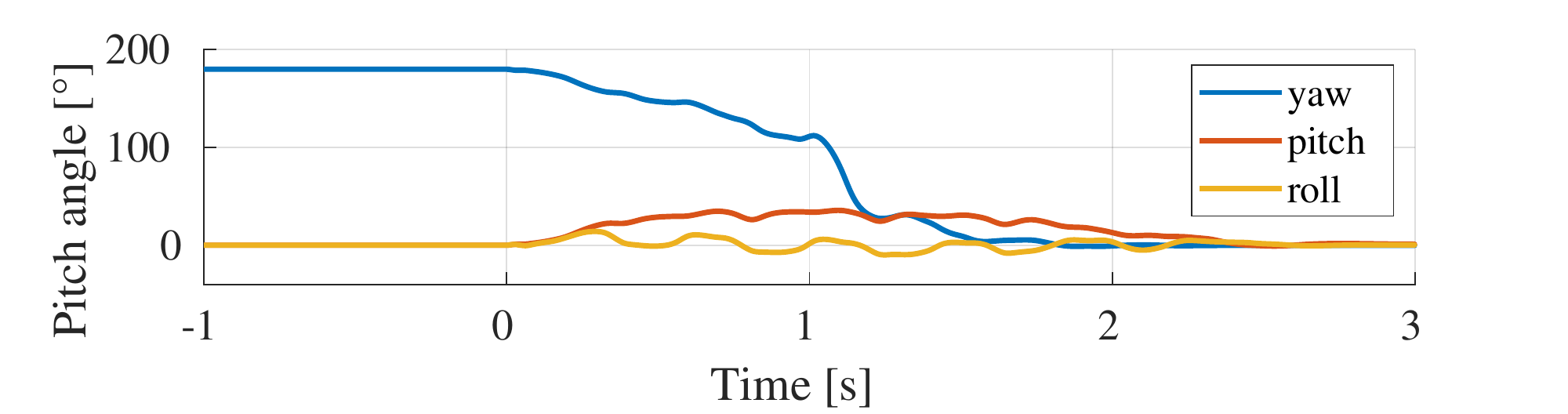}}} 
    \caption{Top: Example of a 3D orientation episode. The robot performs a flip on the yaw axis. Bottom: Evolution of yaw, pitch, and roll during this episode.}%
    \label{fig_3d_orient}
   \vspace{-4mm}
\end{figure}
\begin{figure}[!tb]
    \centering
    \includegraphics[width=\linewidth, trim={0cm, 0, 0cm, 0},clip]{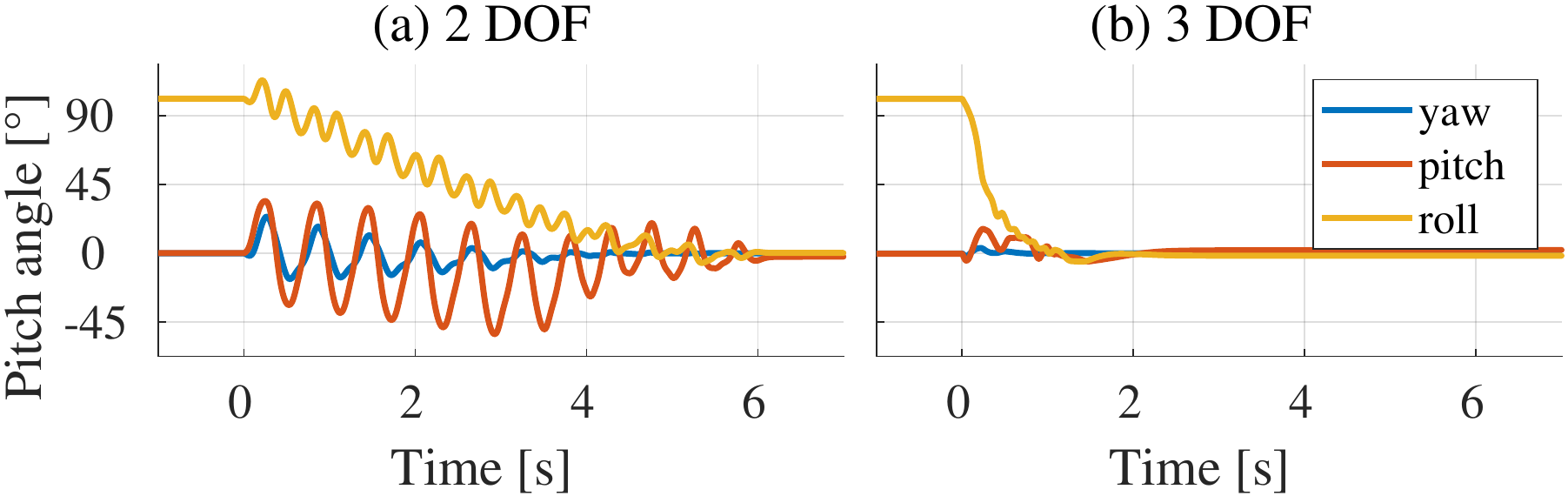}
    \caption{Comparison between  policies reorienting the robot with the original 2 DOF leg and with the added 3\textsuperscript{rd} actuator for hip abduction/adduction. The added degree of freedom allows much faster reorientation.}
    \label{fig_3dof_vs_2dof}
    % \vspace{-5mm}
\end{figure}

With the added actuators, we obtain a policy that successfully controls all axis and can perform any rotation in less than \SI{3}{s}. Figure \ref{fig_3d_orient} provides an example episode with corresponding plots of yaw, pitch, and roll angles.
Furthermore, we evaluate the added performance of the extra actuator by comparing this policy with one trained on the original leg configuration. \\
Figure \ref{fig_3dof_vs_2dof} shows that while legs with two co-planar DOF each are sufficient to control all axis, the added actuator allows the policy to reorient the robot four times faster, which drastically increases agility.
%
%%%%%%%%%%%%%%%%%%%%%%%%%%%%%%%%%%%%%%%%%%%%%%%%%%%%%%%%%%%%%%%%%%%%%%%%%%%%%%%%%%%%%%%%%%%%%%%%%%%%%%%%%%%%%%%%%%%%%%%%%%%%%%%%%%%%%%%%%%%%%%%%%%%%%%%%%%%%%%%%%%%%%%%%%%%%%%%%%%%%%%%%%%%%%%%%%%%%%%
\section{3D Landing}\label{sec_3d_land}
Finally, our most complex task represents landing on a low-gravity celestial body. Similarly to 2D jumping, the robot must reorient itself from a random orientation before landing, but this time it must stay on the ground instead of jumping back.
\subsubsection{Observation space}
Observations are identical to the 2D jumping task, where we use the quaternion representation to extend the orientation to 3D.
\subsubsection{Action space}
As for the 3D attitude control task, all legs are controlled independently, and the extra actuators are added to provide hip abduction/deduction.
\subsubsection{Reward}
We keep the reward terms of Equation \ref{rew_weighted_sum_jump}, but use  $r_{\textit{orient,3D}}$ and set the target velocity $\mathbf{v}_b^*$ of $r_{\textit{jump}}$ to zero. 
\subsection{Training}
Even though we still need to apply a slow curriculum starting with $\eta=0.25$, we see fewer reward oscillations in the landing task than in the orientation one. If the policy fails to reorient the robot or turns in the wrong direction, it will crash on the ground. This creates a strong incentive that seems useful to learn correct reorientation.
\subsection{Simulation performance}
During the fall, our new policy has a similar behavior to the 3D attitude control task.

\begin{figure}[!tb]
    \centering
    \subfloat[][]{{\includegraphics[trim={0 5mm 0 0},clip,width=0.35\linewidth]{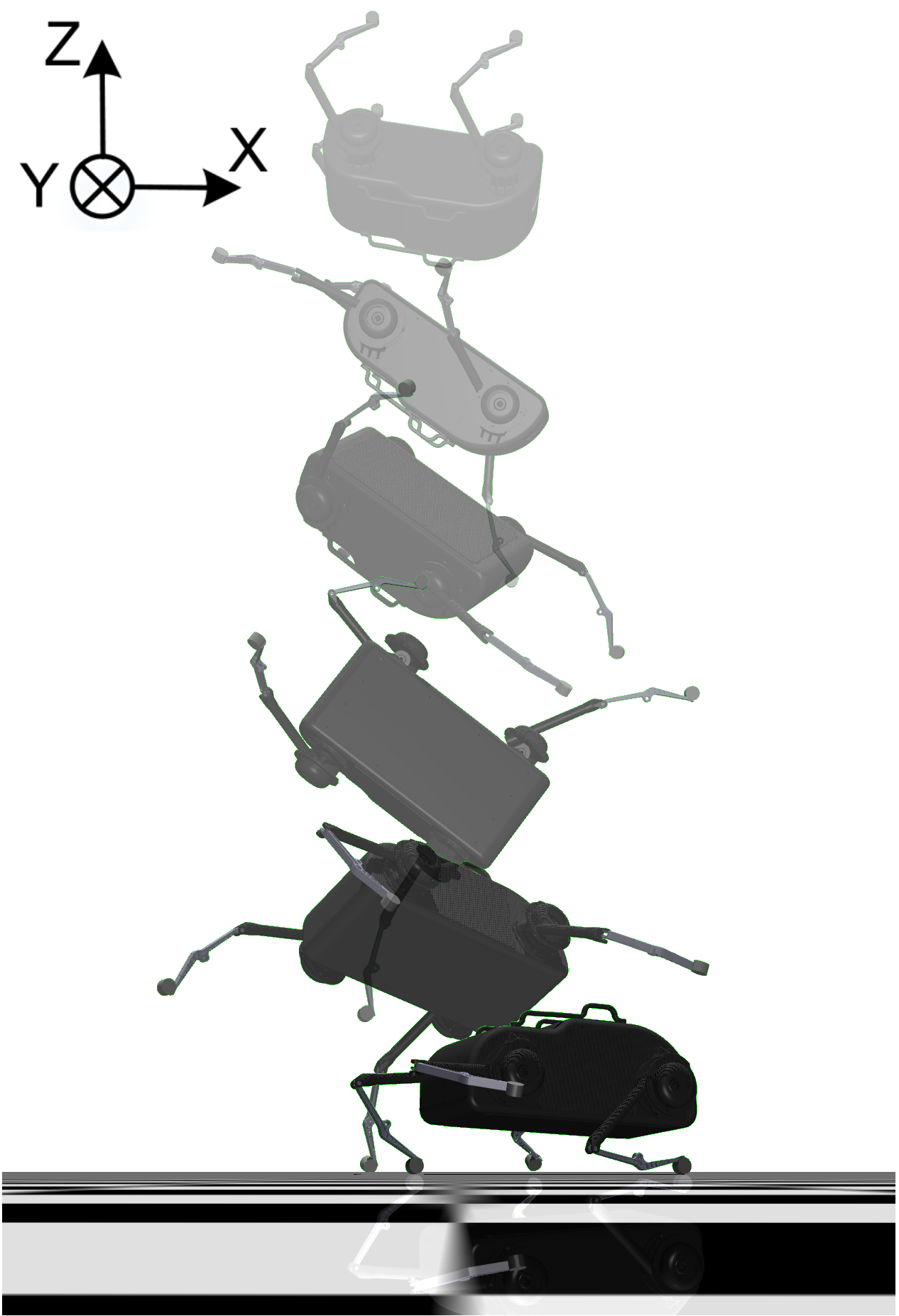}}}
    \,
    \subfloat[][]{{\includegraphics[width=0.35\linewidth]{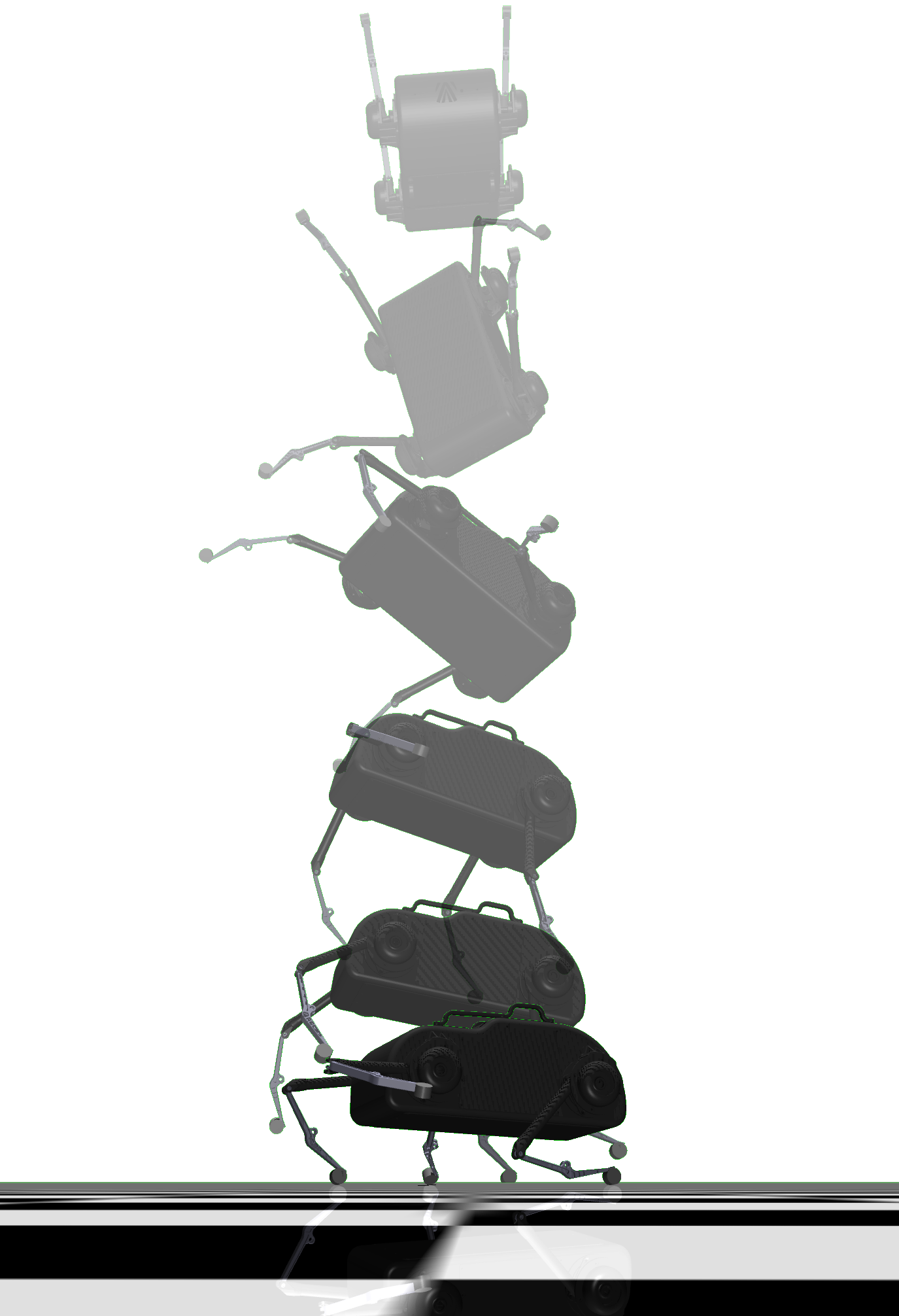}}} 
    
    \caption{Two examples of 3-dimensional landing episodes. The robot starts with $v_b=(0, 0, -1m/s)$ and $\textit{roll}=\pi$
    in episode (a) and $\textit{roll}=\pi, \textit{yaw}=\pi/2$ in episode (b). In both cases the robot has \SI{3}{s} to re-orient itself for landing.}
    \label{fig_land}%
\end{figure}

\begin{figure}[!tb]
  \centering
  \begin{subfigure}{0.45\linewidth}
        \centering
        \includegraphics[width=\linewidth, trim={6cm, 2cm, 6cm, 5cm}, clip]{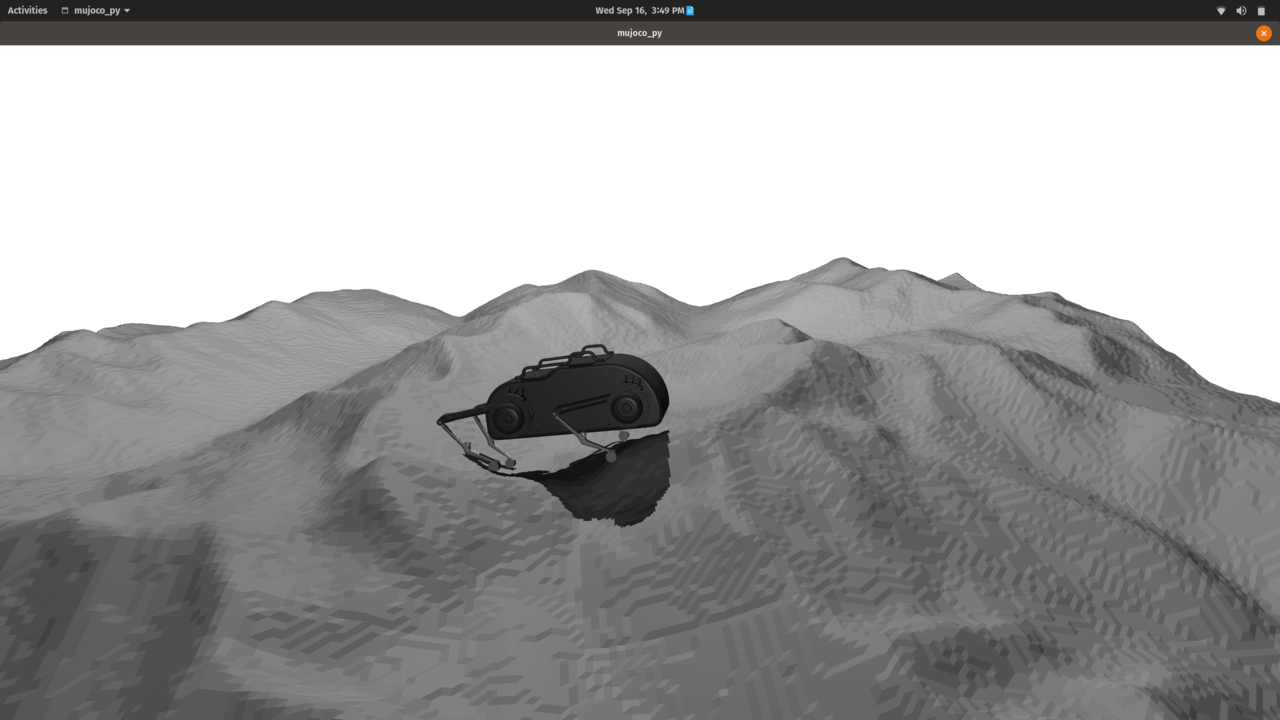}
        \includegraphics[width=\linewidth, trim={6cm, 1cm, 6cm, 6cm}, clip]{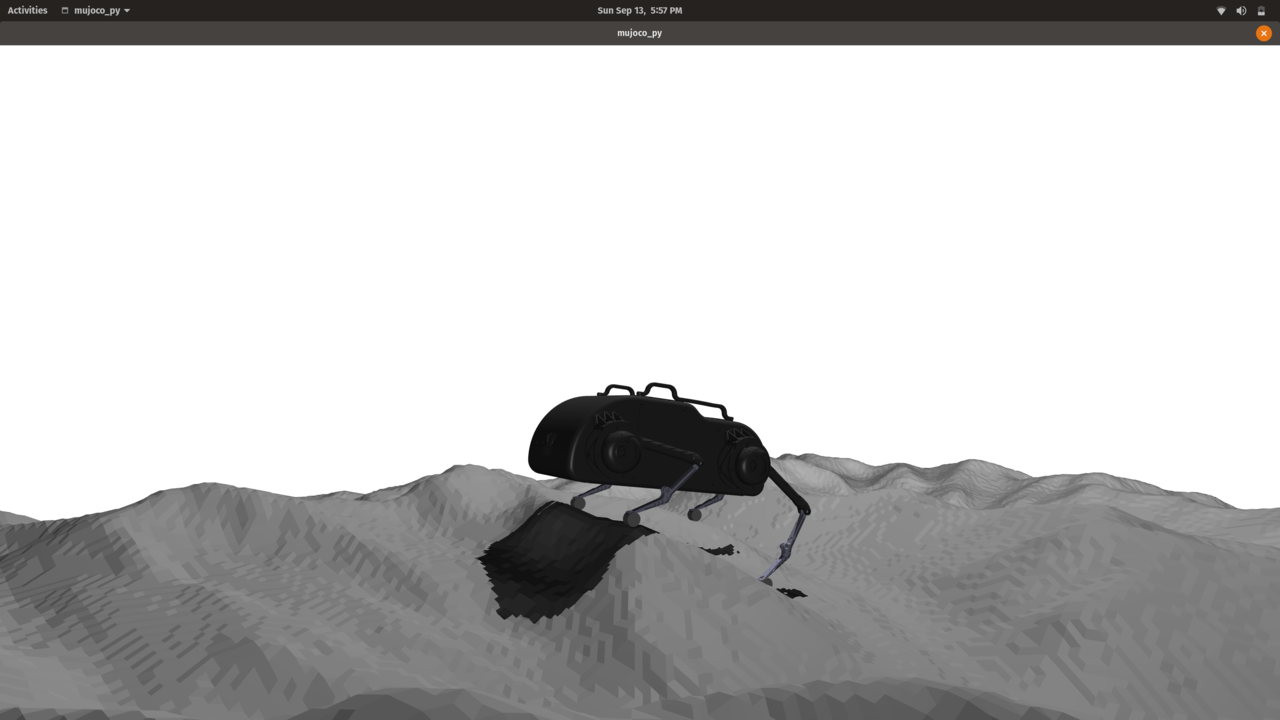}
        \caption{}
    \end{subfigure}
  \begin{subfigure}{0.45\linewidth}
        \centering
        \includegraphics[width=\linewidth, trim={6cm, 2cm, 6cm, 5cm}, clip]{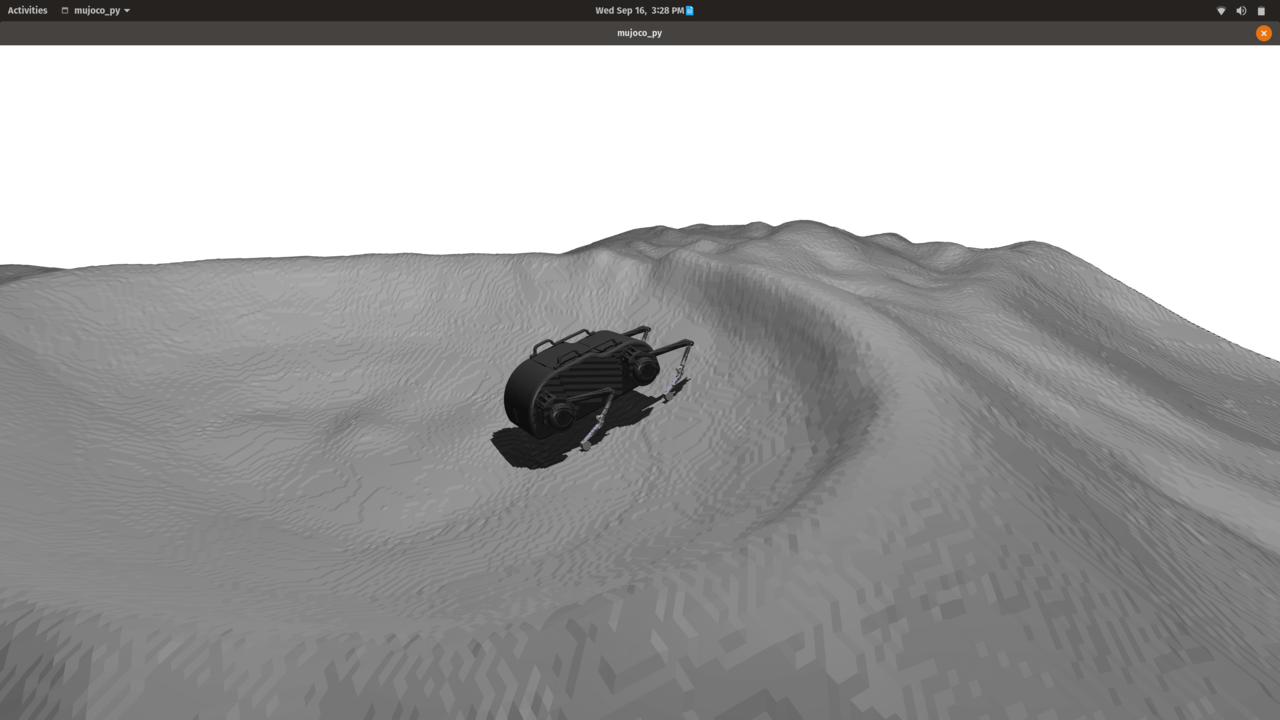}
        \includegraphics[width=\linewidth, trim={6cm, 1cm, 6cm, 6cm}, clip]{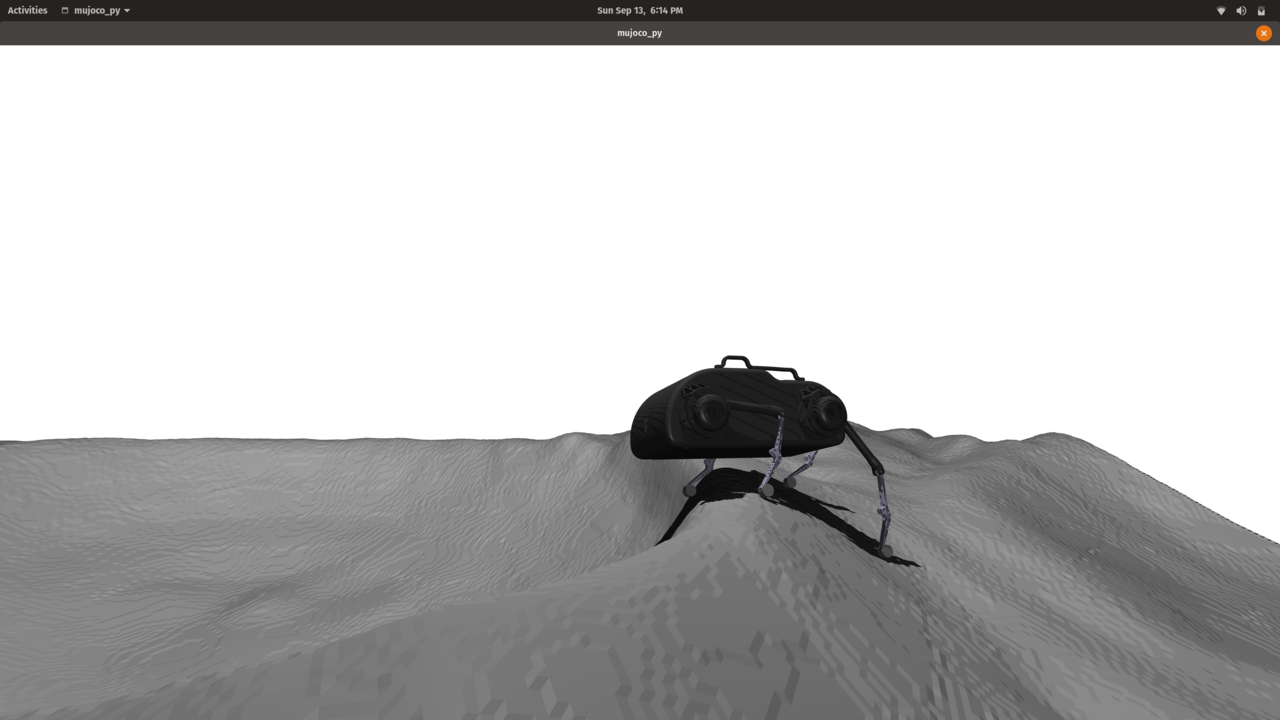}
        \caption{}
    \end{subfigure}
  \caption{The policy is trained on terrain (a) with uniform valleys of depths up to \SI{0.7}{m}, but generalizes well to terrain (b), with a steep crater of \SI{1.5}{m} depth.} 
  \label{fig_land_rough}
  \vspace{-5mm}
\end{figure}

However, instead of matching the target orientation perfectly, higher importance is laid on having the feet in a suitable landing configuration.
This is due to the fact that during each training episode, we limit the flight time to \SI{3}{s} before hitting the ground. As a result, the robot does not always have time to reach an ideal landing orientation, and the policy learns complex mechanisms to recover from these situations. For example, it extends one of the lower legs to push against the ground and quickly reorient the robot right before landing. Figure \ref{fig_land} provides two example episodes.
After these promising results, we investigate our policies' capability to land on uneven terrain resembling an asteroid landscape. It seems reasonable to assume that in order to achieve robust landing capacity, the policy must be able to perceive a representation of the terrain. Nevertheless, we obtain surprisingly positive results even without providing any such measurements. Without modifying the observations, actions, or rewards, we train a new policy able to land on various uneven terrains. Figure \ref{fig_land_rough} shows the final position of four landings on such terrains.
%%%%%%%%%%%%%%%%%%%%%%%%%%%%%%%%%%%%%%%%%%%%%%%%%%%%%%%%%%%%%%%%%%%%%%%%%%%%%%%%%%%%%%%%%%%%%%%%%%%%%%%%%%%%%%%%%%%%%%%%%%%%%%%%%%%%%%%%%%%%%%%%%%%%%%%%%%%%%%%%%%%%%%%%%%%%%%%%%%%%%%%%%%%%%%%%%%%%%%
\section{Discussion}
\label{sec_discussion}

Having demonstrated the capabilities of model-free learning methods, we now discuss important general aspects of our approach.

\subsection{Training process analysis}
\begin{figure}[!tb]
  \centering
    \begin{subfigure}{0.49\linewidth}
        \centering
        \includegraphics[width=\linewidth, trim={4cm, 10.5cm, 5cm, 11cm},clip]{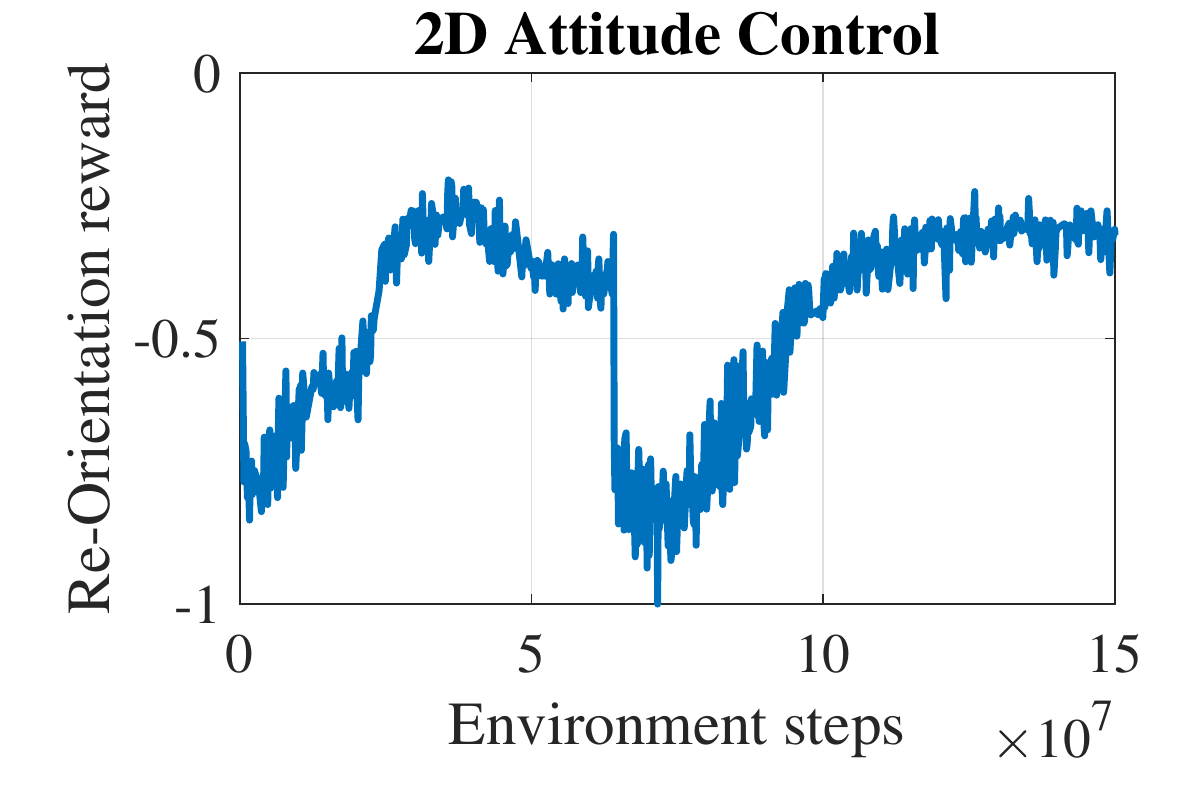}
        \includegraphics[width=\linewidth, trim={4cm, 10.5cm, 5cm, 11cm},clip]{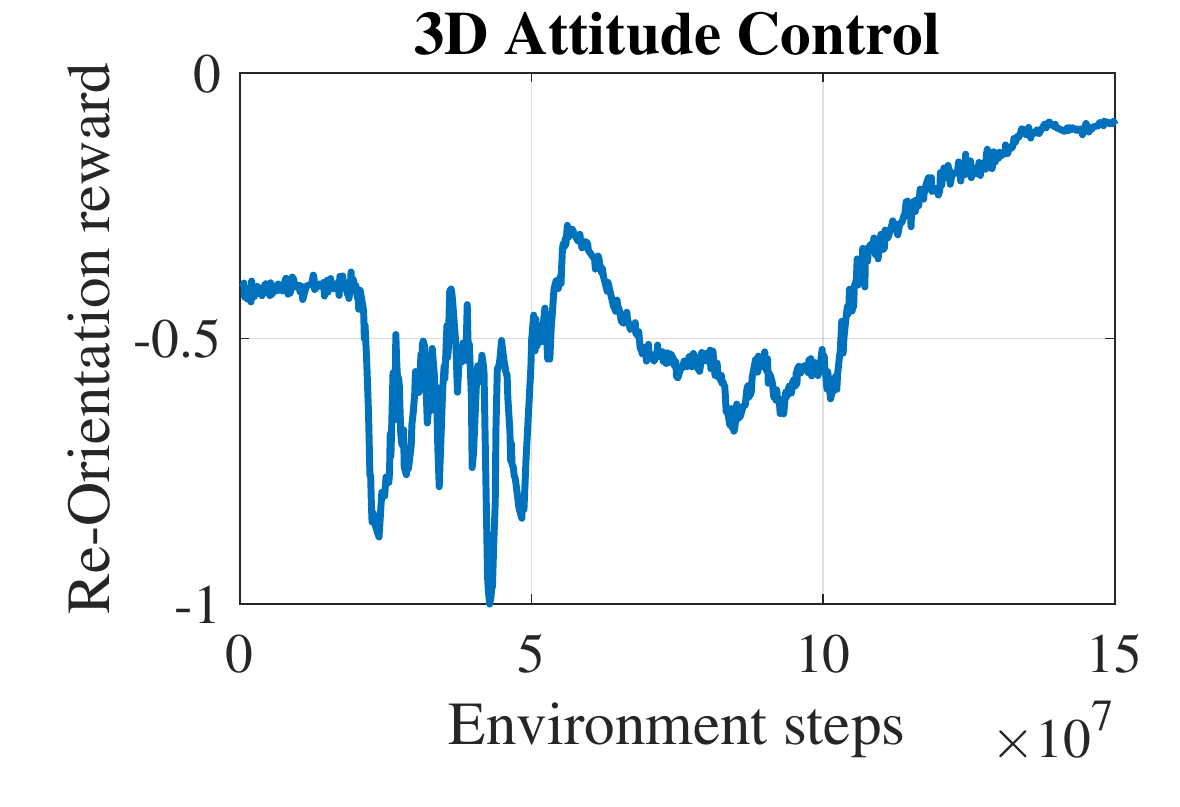}
    \end{subfigure}
    \begin{subfigure}{0.49\linewidth}
        \centering
        \includegraphics[width=\linewidth, trim={4cm, 10.5cm, 5cm, 11cm},clip]{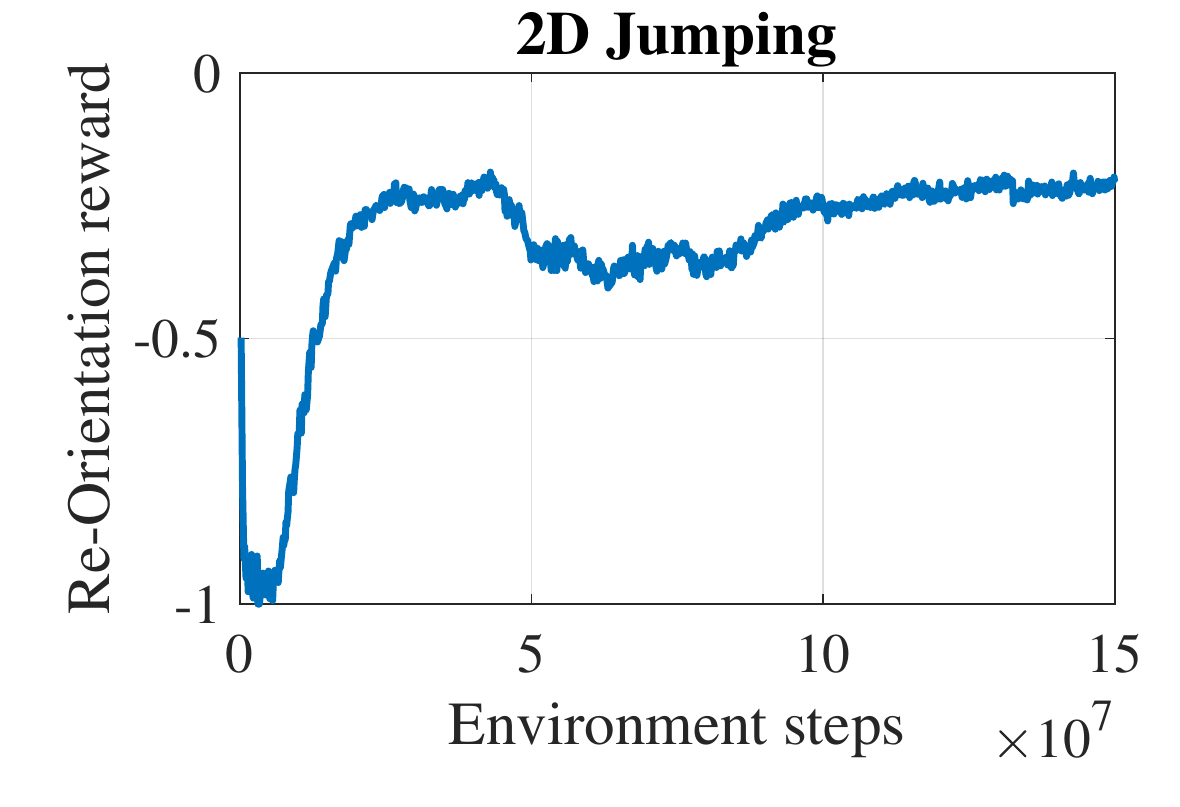}
        \includegraphics[width=\linewidth,trim={4cm, 10.5cm, 5cm, 11cm},clip]{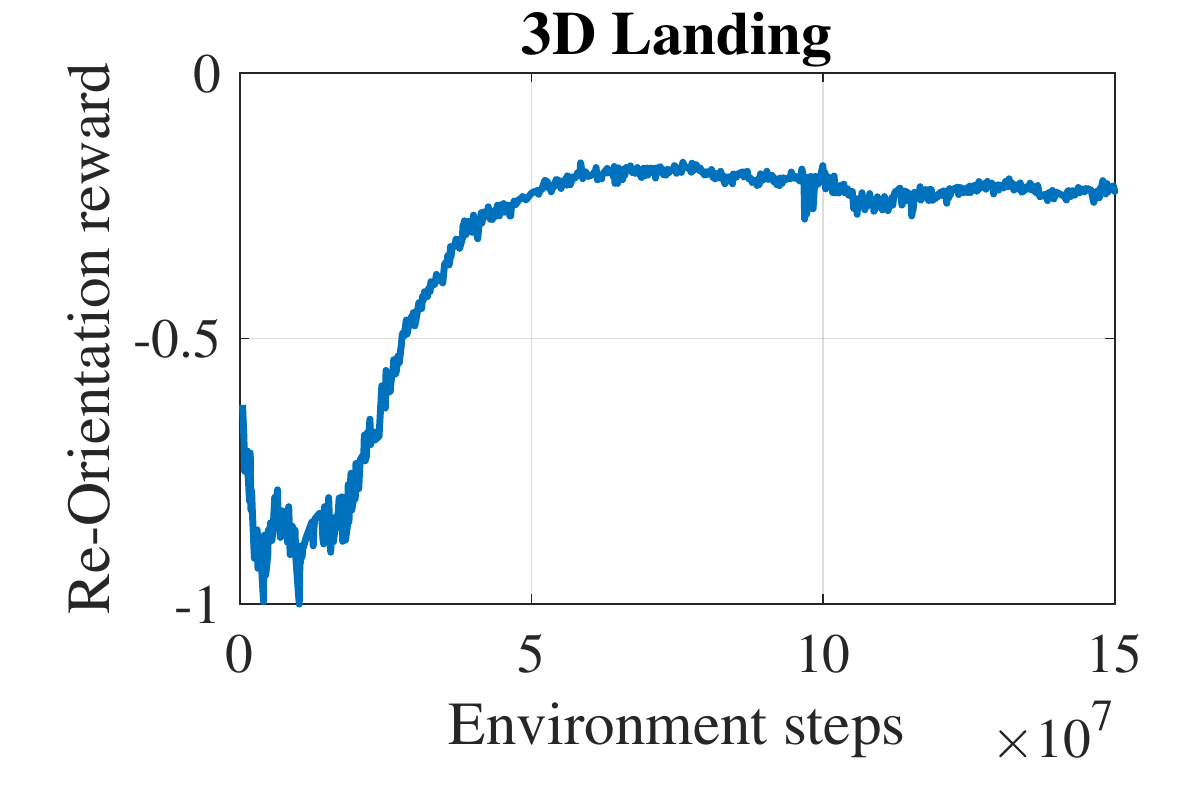}
    \end{subfigure}
  \caption{Training runs of each task, showing instabilities in the training process. The absolute values of the rewards are not comparable, because of differences in scaling and initial state distributions, but their evolution can be analyzed.} 
  \label{fig_rewards}
\end{figure}

As mentioned before, the training process can be unstable. Figure \ref{fig_rewards} shows the evolution of re-orientation rewards throughout the training process. For the 2D attitude control task, we see that the policy diverges from its solution after a positive progression, and the reward drops abruptly. The policy recovers and finds a suitable solution again, but preventing these drops could significantly reduce the required training steps. We use a curriculum on the initial pitch angle distribution during training, which results in a slow decline of the reward, but it does not explain the sharp drop.

In the 3D re-orientation case, the problem is even more severe. We see an oscillation of the reward for up to 60 million steps, and the policy does not find a solution until 100 million steps (corresponding to 4 hours of computation time). These oscillations are not merely random noise since many policy updates constitute each peak. We can also see that throughout this oscillatory regime, the reward stays below the initial value of $-0.4$, which corresponds to an agent doing nothing or shaking its legs randomly. We confirm this by visualizing the behavior of a policy obtaining the lowest reward and find that it learns to use the feet to rotate the robot's base, but does so in the wrong direction. This might be related to the fact that all rotational degrees of freedom are highly coupled, and it is hard to act on one of them without affecting the others. After this long period, we finally see that the policy finds a solution and the reward steadily increases without further instabilities.

Interestingly, we do not see these drops and oscillations in the jumping and landing tasks\footnote{The slight drop in the 2D jumping task is caused by the curriculum slowly increasing the task complexity.}. In these cases, when the robot is not correctly oriented, it crashes into the ground, causing a high penalty. This sparse reward provides a strong incentive to re-orient the robot correctly and seems beneficial for a smooth training process.

\subsection{On-board state estimation}
In order to solve our tasks, the robot must be able to measure its orientation and the distance to the ground. In this work, we focused on locomotion control aspect and assumed that these measurements are readily available by the state estimator. When conducting tests on the physical robot, we use a Vicon system to track the robot's absolute position and orientation with sub-millimeter accuracy while fixing the position of the walls. In a more realistic scenario, the robot would need to estimate those quantities using on-board sensors. A possible solution could is the fusion of inertial measurements with exteroceptive sensors, such as range sensors placed around the robot's body. Similar methods have been developed for safely landing spacecrafts on planets \cite{johnson2007design} or perform on-orbit pose estimation between two satellites \cite{opromolla2017onorbitreview}. Nevertheless, the development of sensors and state estimation techniques is an active area of research and highly depending on the target environment (lightning conditions, mass/power constraints, available computational power, etc.) and thus considered out of scope for this work.

\subsection{Terrain shape}
In our current tasks, the robot lands on perfectly flat and rigid terrain. Even though the results of Sec. \ref{sec_3d_land} show that we can train a policy to land blindly on uneven terrain, in order to achieve better performance and robustness, the policy should receive a representation of the terrain as part of its observation. Reinforcement learning has the advantage of scaling well to increasing task complexity, and it is not limited in any way to the type of observations it receives \cite{RubiksCube}.

\subsection{Comparison to Model-Based Approaches}
As stated above, to the best of our knowledge, model-based approaches would struggle to solve our task end-to-end. However, sub-parts could be solved individually:

\textit{2D and 3D re-orientation} can be achieved by selecting hand-tuned trajectories, such as a circular motion of the feet. However, such a solution would be far from optimal in terms of re-orientation time. An optimization approach could be used, but a free-floating system is surprisingly difficult to control with such methods due to challenges in both kinematics and dynamics that are not present in earth-based systems \cite{PapadopoulosFreeFloating, PapadopoulosNonHolonomic}.

\textit{Ground contact dynamics} can be solved with model-based methods. These either require many assumptions, such as fixed gaits and contact patterns \cite{FFmpc} or require a computationally intensive trajectory optimization, which is hard to solve in real-time \cite{NeunertTrajOpt}.

Overall, even though model-based techniques could be applied by combining the solution for landing and jumping. We argue that it would require an extensive amount of effort to tune each case and that the combination would not exploit the full potential of the robot compared to an end-to-end solution. For example, emerging behavior such as utilizing ground contacts for last-second re-orientation corrections would be missed by splitting the phases. A learned approach does not restrict possible solutions, solves the complete problem, and can be easily computed in real-time on the robot.
\\
A lack of thorough verification, which is of particular importance for space systems, is the main criticism of non-deterministic, learning-based approaches. However, we reduce the problem during both training and deployment. Throughout the training, we include safety margins on collisions and penalize risky high-frequency motions. Once deployed, we show that we can include hard constraints on joint positions and torques to satisfy the system's hardware limitations, allowing us to deploy our policies on the physical robot safely.

\section{Conclusions and future work}
\label{sec_conclusion}

Using an off-the-shelf deep reinforcement learning algorithm and sim-to-real transfer via domain randomization, we have shown that a model-free learning approach solves multiple locomotion tasks, which, as of now, do not have other available solutions. In our application, we solve the task of attitude control during an extensive flight-phase, smooth landing, and take-off of a quadrupedal robot by solely using its limbs.
Our method significantly reduces the weight and mechanical complexity compared to a similar implementation relying on traditionally used reaction wheels. 
The neural network handles the highly non-linear dynamics and uses non-trivial solutions to solve the tasks. With the learned policy, the robot avoids self-collisions, handles the discontinuities invoked by hard contacts, and performs aggressive motions to control its attitude rapidly. For example, the learned motion allows for (pitch-)orientation changes of 90° in less than \SI{2.5}{s} in simulation and on the real robot. Furthermore, it solves the tasks end-to-end, without using predefined trajectories, and is free to design any acyclic, non-continuous solution, or even exploit the terrain. Once trained, the neural network's inference requires minimal time to compute, performs immediate actions, and allows a practical deployment on physical hardware. We have validated two-dimensional tasks with extensive tests on the physical robot, where we demonstrated several consecutive jumps without human intervention. The same approach readily scales to three dimensions and, in simulation, also handles arbitrarily shaped terrain.

As part of future extensions, we plan to work towards deploying 3D policies in the real world. This requires a robot with three degrees of freedom per leg, and the need to replicate a low-gravity environment. The first part can easily be solved by switching to another quadruped robot and is not expected to significantly change the approach. A microgravity environment can be obtained on a parabolic flight for approximately \SI{20}{s} at a time.
These flights have been previously used to test a robotic arm \cite{FlightArm} and even test attitude control of a small robot with an attached manipulator \cite{Flight3DRobot}. In this work, we focused exclusively on a quadruped robot, but the approach could be extended to the control of satellites with robotic arms.
 
\section{ACKNOWLEDGMENT}
\label{sec_acknowledgment}
The authors would like to thank Martin Werner Zwick and Gianfranco Visentin from the Automation and Robotics section of the European Space Agency for their organizational efforts in realizing the tests. 

\bibliographystyle{IEEEtran}
\balance
\bibliography{bibtex}
\addcontentsline{toc}{chapter}{Bibliography}
\end{document}